\documentclass[11pt]{article}

\usepackage[final]{acl}

\usepackage{times}
\usepackage{latexsym}

\usepackage[T1]{fontenc}

\usepackage[utf8]{inputenc}

\usepackage{microtype}

\usepackage{inconsolata}

\usepackage{graphicx}
\usepackage{amsmath}
\usepackage{booktabs}
\usepackage{textcomp}
\usepackage{svg}
\usepackage{enumitem}
\usepackage{float}

\title{Where Animacy Lives in Large Language Models:\\ Tracing the Circuits of the Animacy Concept}

\author{
 \textbf{Samuele Punzo\textsuperscript{1}}, 
 \textbf{Giovanni Cinà\textsuperscript{2,3*}}, 
 \textbf{Sandro Pezzelle\textsuperscript{3*}}
\\
\\
 \textsuperscript{1} Graduate School of Informatics, University of Amsterdam \\
 \textsuperscript{2} Department of Medical Informatics, Amsterdam University Medical Center \\
 \textsuperscript{3} ILLC, University of Amsterdam \\
 \small{
    \textsuperscript{*}Equal supervision
}
\\
Corresponding author: \href{mailto:email@domain}{samuele.punzo@student.uva.nl}
}

\begin{document}
\maketitle

\begin{figure*}
    \centering
    \includegraphics[width=1\linewidth]{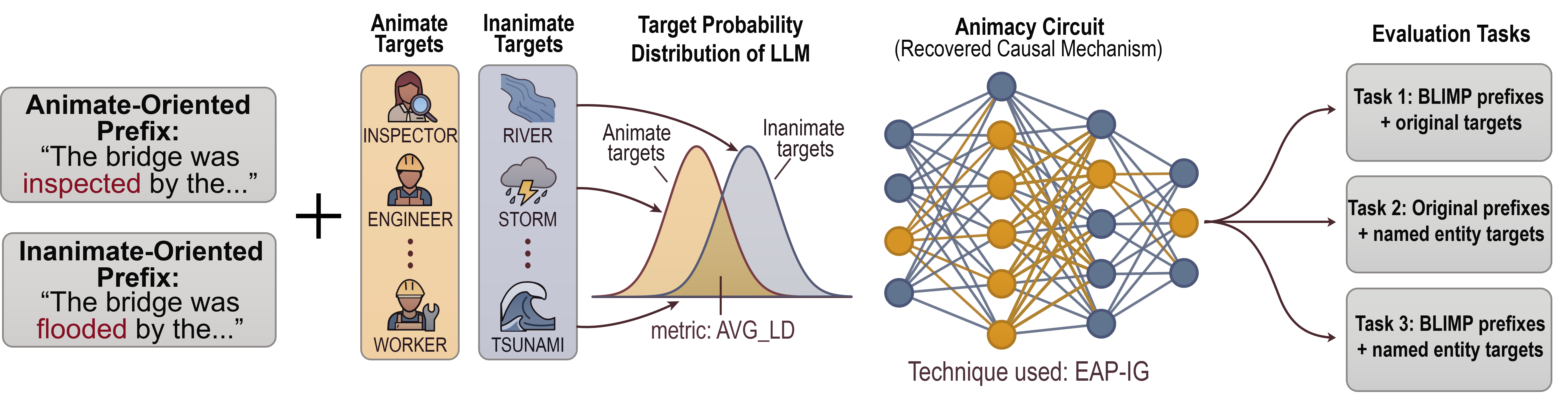}
    \caption{Circuit discovery and evaluation pipeline. AVG\_LD is the average logit-difference metric used to define the task.}
    \label{fig:pipeline}
\end{figure*}

\begin{abstract}
Distinguishing animate from inanimate concepts in written language requires more than shallow text processing, as it involves recognizing complex selectional constraints and contextual cues, such as verb-argument interactions. Yet, current large language models (LLMs) appear to be capable of doing it. We investigate whether this animacy-sensitive behavior of LLMs can be traced to a localized set of causally relevant components and connections. To do so, we construct a controlled dataset of minimal pairs and perform circuit discovery on four open-weight models. Through in-depth experiments and ablations, we show that a causal mechanism responsible for handling animacy in these models does exist, thus discovering an \textit{animacy circuit}. At the same time, this circuit appears to be less localized compared to other known ones and generalizes only partially across models and animacy tasks, confirming the distributed, context-dependent, and somewhat graded nature of the animacy concept.
\end{abstract}

\section{Introduction}
Large Language Models (LLMs) exhibit impressive capabilities on a wide range of linguistic phenomena, raising the question of whether they also capture complex conceptual distinctions beyond the language surface. Unlike grammatical tasks, animacy---whether an entity is alive, sentient, or capable of intentional action~\citep{comrie-animacy}---is often not signalled by any visible feature. Yet, animacy affects syntactic structure, as animate entities are often preferred in subject and agent positions~\citep{caplan1994interaction}; thematic-role assignment, as in \textit{``The dog chased the ball''}, where the animate noun is a more plausible agent~\citep{MACWHINNEY1984127}; selectional preferences, as in \textit{``The surgeon examined the patient''}~\citep{PACZYNSKI2012426}; and co-reference, as in \textit{``The dog chased the ball because it was red''}, where \textit{it} naturally refers to the inanimate noun~\citep{Bader21102023}. 

An ideal scenario for testing animacy understanding is, therefore, sentence completion. Predicting whether an unfinished sentence should be continued with an animate or inanimate concept requires models to understand complex selectional constraints and contextual cues, such as verb-argument interactions. For example, the prefix \textit{``The bridge was inspected by the''} is likely followed by an animate noun, such as \textit{``engineer''}, whereas \textit{``The bridge was flooded by the''} favors an inanimate continuation, i.e., the noun \textit{``storm''}. Previous studies employed this and similar experimental paradigms to study animacy comprehension in LLMs, mostly focusing on behavioral evaluation~\citep{coll-ardanuy-etal-2020-living} and surprisal- or representation-based analyses~\citep{hanna2023language, kauf2022event}. While insightful, this body of work has not investigated the \textit{causal mechanisms} underlying the models' handling of animacy comprehension. This leaves an open question: Is there a dedicated set of model components---an \textit{animacy circuit}---causally responsible for this ability?

We address this gap by investigating three main questions: First, is animacy-sensitive completion supported by a localized circuit, that is, a relatively small set of model components and edges? Second, is such a circuit causally sufficient and necessary for the behavior? Third, does the circuit generalize across models and across related variants of the animacy task? 
To study these questions, we construct a controlled dataset of minimal pairs designed to isolate animacy-sensitive continuation preferences \textcolor{black}{in English}, i.e., to assess whether the model prefers an animate or inanimate concept to continue the sentence (see the example above). 
Following modern approaches in mechanistic interpretability, we perform circuit discovery via Edge Attribution Patching with Integrated Gradients~\cite[EAP-IG;][]{hanna2024faithfaithfulnessgoingcircuit} on four open-weight language models: GPT-2 Small, Llama 3.2 3B, Gemma 3 4B, and Qwen 3 4B~\citep{gemmateam2025gemma3technicalreport, yang2025qwen3technicalreport, grattafiori2024llama3herdmodels, Radford2019LanguageMA}. \textcolor{black}{\autoref{fig:pipeline} summarizes the full workflow, from dataset construction and metric-based filtering to circuit discovery and transfer evaluation.}


For each of these models, we identify a circuit for animacy that is both sufficient and necessary to recover the behavior on the task. However, our analyses show that these circuits should not be interpreted as universal animacy circuits. Indeed, their transfer to related animacy settings is uneven across datasets and models, suggesting that they are partly constrained by the specific task and data. These results point to a more nuanced picture: animacy-sensitive behavior is mechanistically structured and can be partially localized inside language models, but animacy remains a complex semantic phenomenon that might not be captured by a single circuit generalizing uniformly across settings. Datasets and code are available at: \href{https://github.com/SamuelePunzo/animacy-circuit}{github.com/SamuelePunzo/animacy-circuit}.

\section{Related Work}

\subsection{Animacy in Linguistics and Cognition}
Animacy is a central semantic concept in human cognition and language, distinguishing between entities that are living, agentive, or capable of intentional action and those that are not. It has been studied from both linguistic and cognitive-neuroscientific perspectives. Linguistic work has shown that animacy shapes grammatical structure and sentence interpretation \citep{comrie-animacy, MACWHINNEY1984127, KATHRYNBOCK198547, gregorio-etal-2025-cross}, while cognitive and neuroscientific work suggests that animate and inanimate entities are treated differently in perception, memory, and semantic knowledge \citep{Caramazza1998-xf, New2007-vj, Nairne2013-wf}. The animacy hierarchy \citep{comrie-animacy} captures the tendency for human, animal, and inanimate entities to behave differently across grammatical systems, influencing phenomena such as argument realization, case marking, and word order. More recent cross-linguistic work has further shown that animacy plays a systematic role in grammatical structure across languages \citep{gregorio-etal-2025-cross, li-etal-2026-mechanistic}. Behavioral work has also shown that animacy guides thematic-role assignment and sentence processing: Animate entities are more likely to be interpreted as agents and are often processed more easily in subject or first-mentioned positions \citep{MACWHINNEY1984127, KATHRYNBOCK198547, TRAXLER200269}.

\subsection{Animacy in LLMs}

Unlike many other phenomena, animacy-sensitive behavior cannot be solved from local grammatical and lexical cues alone. This requires models to use contextual semantic information effectively.
Earlier studies on word embeddings found that animacy is a difficult semantic property to recover reliably from purely distributional representations~\citep{lucy-gauthier-2017-distributional}. More recent work by \citet{hanna2023language} shows that transformer language models are sensitive to animacy-related selectional constraints, both in typical cases and in contexts where normally inanimate entities are made animate. Other analyses of contextual language models likewise suggest that animacy-related information can be encoded in internal representations \citep{coll-ardanuy-etal-2020-living} and can affect model behavior on syntactic or semantic prediction tasks \citep{kauf2022event}. 

\textcolor{black}{\citet{li-etal-2026-mechanistic} investigated how animacy influences active-passive structure choice in GPT-2 Small, identifying attention heads that mediate this effect. Our work addresses a complementary question: whether a localized internal mechanism underlies the animacy-completion behavior in LLMs.}

\subsection{Circuit discovery}

Recent work in mechanistic interpretability has developed a range of techniques for studying how language models implement particular behaviors internally. Rather than treating models only through their input-output behavior, these methods analyze the role of their components and the connections between them \citep{olah2020zoom, elhage2021mathematical}. These techniques have been applied to several NLP-relevant phenomena, including indirect object identification (IoI) \citep{wang2022interpretabilitywildcircuitindirect}, factual association storage and editing \citep{meng2023locatingeditingfactualassociations}, feed-forward key-value memory mechanisms \citep{geva2021transformer}, and numerical comparison \citep{greater-than}.

One approach within this framework is circuit discovery. A circuit is a subgraph of model components and connections responsible for a target behavior \citep{olah2020zoom, elhage2021mathematical}. In Transformer language models, circuit discovery aims to identify the attention heads, MLPs, or directed edges whose activity causally contributes to a given output. This is typically achieved through interventions on activations or edges. Activation patching replaces activations from one clean input with those from a corrupted counterpart and measures the resulting effect on the output \citep{wang2022interpretabilitywildcircuitindirect,conmy2023automatedcircuitdiscoverymechanistic}, but it can be computationally expensive. More scalable alternatives approximate patching with gradients: attribution patching uses a first-order Taylor approximation \citep{Nanda_EAP}, while Edge Attribution Patching (EAP) applies this idea to directed connections between components \citep{syed-etal-2024-attribution}. More recent methods improve the faithfulness of these scores through integrated gradients in EAP-IG \citep{hanna2024faithfaithfulnessgoingcircuit} or adaptive integration paths that avoid saturation effects \citep{zhang2026eap}. Recent work also emphasizes that the reliability of a discovered circuit depends on how it is evaluated. Circuits recovered by patching-based methods can depend strongly on the dataset, task metric, and clean--corrupt contrast used to define the behavior \citep{mueller2024missedcausesambiguouseffects, zhang2024bestpracticesactivationpatching, formal-hanna-etal-2026}. High faithfulness in a single setting therefore does not necessarily imply a robust or task-general mechanism. Building on this, we use circuit discovery as our first step: After identifying candidate animacy circuits with EAP-IG,\footnote{EAP-GP is reported to be approximately five times slower than EAP-IG, making it less suitable for our extensive cross-model analyses} we evaluate them through sufficiency, necessity, random-edge controls, stability across discovery samples, and transfer to related task variants.

\section{Methods}

\subsection{Data}
\label{ssec:dataset_creation}

To investigate whether models represent animate vs.~inanimate concepts through a dedicated mechanism, we need a dataset offering a controlled contrast between the two types, while ensuring semantic plausibility. Existing datasets do not fully satisfy these requirements. For example, BLiMP \textit{subject\_passive}~\citep{warstadt2020blimp} contains sentence pairs such as \textit{``Amanda was respected by some waitresses''} vs.~\textit{``Amanda was respected by some picture''}, where the animate contrast appears only in the final token. This is incompatible with activation/attribution patching: if we forward the complete sentences, the contrastive noun is already present in the input, so there is no target whose probability we can compare. We would therefore have to use the shared prefix, \textit{``Amanda was respected by some''}, but this prefix is identical across the two BLiMP items, meaning that the clean and corrupted inputs would have no animate contrast to patch. Similarly, \citet{gregorio-etal-2025-cross} provide semantically meaningful sentences spanning multiple degrees of animacy, hierarchically ordered as follows: Human \textrightarrow Imaginary being \textrightarrow Animal \textrightarrow Plant/bacteria \textrightarrow Inanimate, but it does not offer a minimal-pair structure suitable for circuit discovery. For this reason, we built a new dataset designed to provide both a strong animate--inanimate contrast and semantically plausible prefixes. We use the template:
\begin{quote}
``The [\textit{patient}] was [\textit{verb p.p.}] by the''
\end{quote}
where the \textit{patient} and \textit{past participle verb} are filled following the procedure below.

To create semantically valid sentences, we ask an LLM~\cite[GPT 5.4;][]{GPT5.4} to generate joint sets of patients and verbs (henceforth referred to as semantic frames) for several semantic categories. We then manually inspect each frame to verify its correctness and consistency; subsequently, we fill the sentence template frame by frame (see Appendix~\ref{sec:dataset_building} for details). For each frame, we instantiate the template with every patients and every pair from \textit{animate\_verbs} $\times$\textit{inanimate\_verbs}. This yields two matched prefixes, where \(p\) is the patient noun and \(v_{an},v_{in}\) are the animate and inanimate verbs:
\[
\begin{array}{lll}
\textsc{clean}: & ``\textit{The } \{p\} \textit{ was} \;\{v_{an}\} \textit{ by the}\text{''} \\
\textsc{corrupt}: & ``\textit{The } \{p\} \textit{ was} \;\{v_{in}\} \textit{ by the}\text{''}

\end{array}
\]
Obtaining minimal pairs like: ``\textit{The victim was examined by the}'' vs. ``\textit{The victim was blinded by the}''.
This procedure yields 20,000 minimal pairs, corresponding to 40,000 prefixes: 20,000 animate-oriented and 20,000 inanimate-oriented. We then score all prefixes for semantic plausibility using Llama 3.3 70B with a zero-shot Yes/No prompt \textcolor{black}{For each clean and corrupt prefix, we compute the probability assigned to the \texttt{Yes} token and use it as a plausibility score. We retain a minimal pair only if both its clean and corrupt prefixes receive a plausibility score above 0.90} (further details in Appendix~\ref{sec:dataset_building}). 
Overall, 16,305 out of 20,000 scored minimal pairs were retained (\(81.5\%\)). From now on, we refer to this set of minimal pairs as our \textbf{animacy dataset}.

\begin{figure*}[t]
    \centering
    \includegraphics[width=0.9\linewidth]{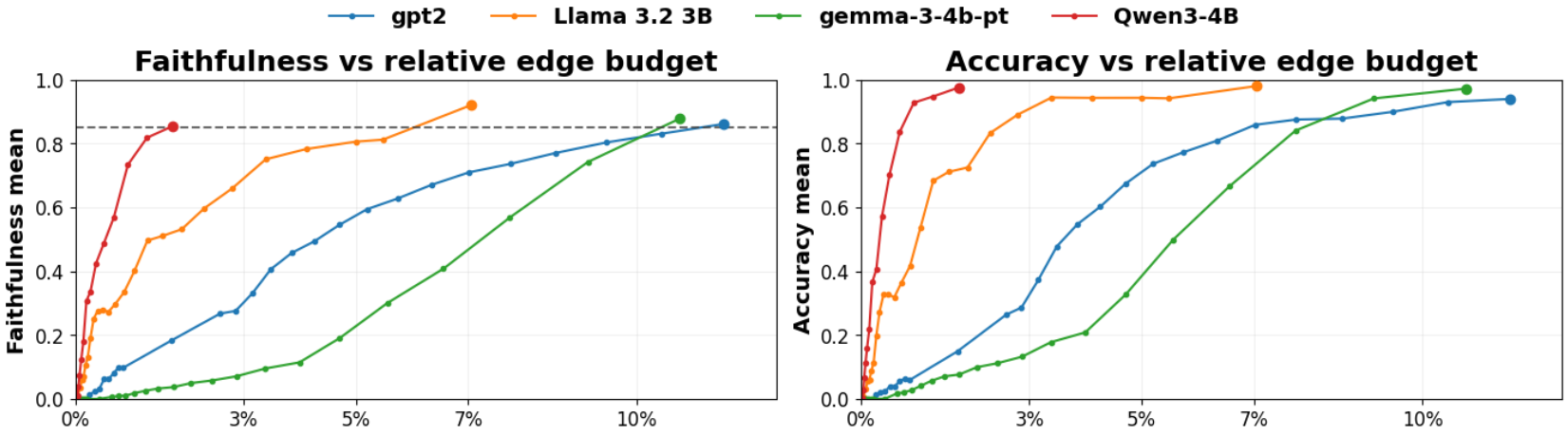}
    \caption{Circuit discovery sweeps for each tested model over collapsed edge budget (\% of ranked model edges). Best viewed in color.}
    \label{fig:circuit_discovery_sweeps}
\end{figure*}

After obtaining our animacy dataset, we construct broad animate and inanimate target sets ($\sim$500 nouns each) designed to mitigate possible biases of specific targets towards the prefixes,\footnote{\textcolor{black}{The prefix "The ship was maneuvered by the" might induce a bias over the logits of `captain' rather than `lawyer'.}} thereby isolating the animacy signal. Starting from candidate nouns extracted from WordNet \citep{miller-1994-wordnet, wordnet}, we distinguish the two classes using a methodology similar to that of \citet{gregorio-etal-2025-cross}. Animate targets are restricted to unambiguous \texttt{noun.person} words, such as \textit{officer}, \textit{manager}, \textit{captain}, \textit{author}, \textit{farmer}, and \textit{detective}. We focus on person nouns because the relevant contrast in our task is between plausible intentional agents and inanimate causes, rather than biological life more broadly. Inanimate targets are non-person nouns belonging to the WordNet categories \texttt{noun.event}, \texttt{noun.phenomenon}, \texttt{noun.state}, \texttt{noun.artifact}, \texttt{noun.substance}, or \texttt{noun.object}. We retain only non-hyphenated words and apply the model-specific single-token filtering described below. The complete WordNet filtering criteria are provided in Appendix~\ref{app:target-construction}.


In practice, because tokenization varies across models, a further filtering step is necessary before running the experiments. For each model, we tokenize both the animacy dataset and the target sets using its model-specific tokenizer. Then, we only retain prompt pairs whose clean and corrupt prefixes, under all models, have aligned token spans, i.e., both their patient noun and verb are single-token. This leads to a subset of our dataset, shared across all tested models, including 9,532 sentence pairs.\footnote{Note that the number of animate/inanimate targets used for evaluation varies slightly across models. They are 450/450 for GPT-2, 442/449 for Gemma-3-4B-PT, and 401/434 for both Llama-3.2-3B and Qwen3-4B.}

\subsection{Task and Metric}
\label{ssec:task-metric}

Following established practices in mechanistic interpretability \citep{heimersheim2024useinterpretactivationpatching, zhang2024bestpracticesactivationpatching}, we evaluated multiple candidate task metrics before selecting the one used for circuit discovery. The comparison between these metrics is reported in Appendix~\ref{sec:metric_investigation}. In our setting, there is no unique correct next token for a given prefix. Instead, the task is defined over two broad semantic target sets: animate nouns and inanimate nouns (see Section~\ref{ssec:dataset_creation}). A clean prefix should increase the model's preference for animate continuations, whereas the corresponding corrupt prefix should increase its preference for the inanimate ones. For this reason, we use a set-based metric rather than one based on a single ground-truth. Our main task metric is the average logit difference between the animate and inanimate target sets. For a prefix \(x\), we define:

{\small
\begin{equation*}
    avg_{LD}(x) = \frac{1}{|A|}\sum_{a \in A} logit(a \mid x) - \frac{1}{|I|}\sum_{i \in I} logit(i \mid x)
\end{equation*}
}

\noindent where \(A\) is the animate target set and \(I\) is the inanimate target set. A positive value indicates that the model favors animate continuations on average, while a negative value indicates that it favors inanimate continuations. For each minimal pair, we compute this metric separately for the clean and corrupt prefixes. A model is considered successful on a pair when the clean prefix yields a positive score and the corrupt prefix yields a negative score:
\begin{align}
    clean_{avg_{LD}} > 0 \land corrupt_{avg_{LD}} < 0 
\label{eq:metric}
\end{align}

We also define a pair-level logit margin as:
\[
margin = clean_{avg_{LD}} - corrupt_{avg_{LD}}
\]

This margin measures how strongly the model separates the clean and corrupt sentences in the intended direction.


\begin{figure}[t]
    \centering
    \includegraphics[width=0.87\linewidth]{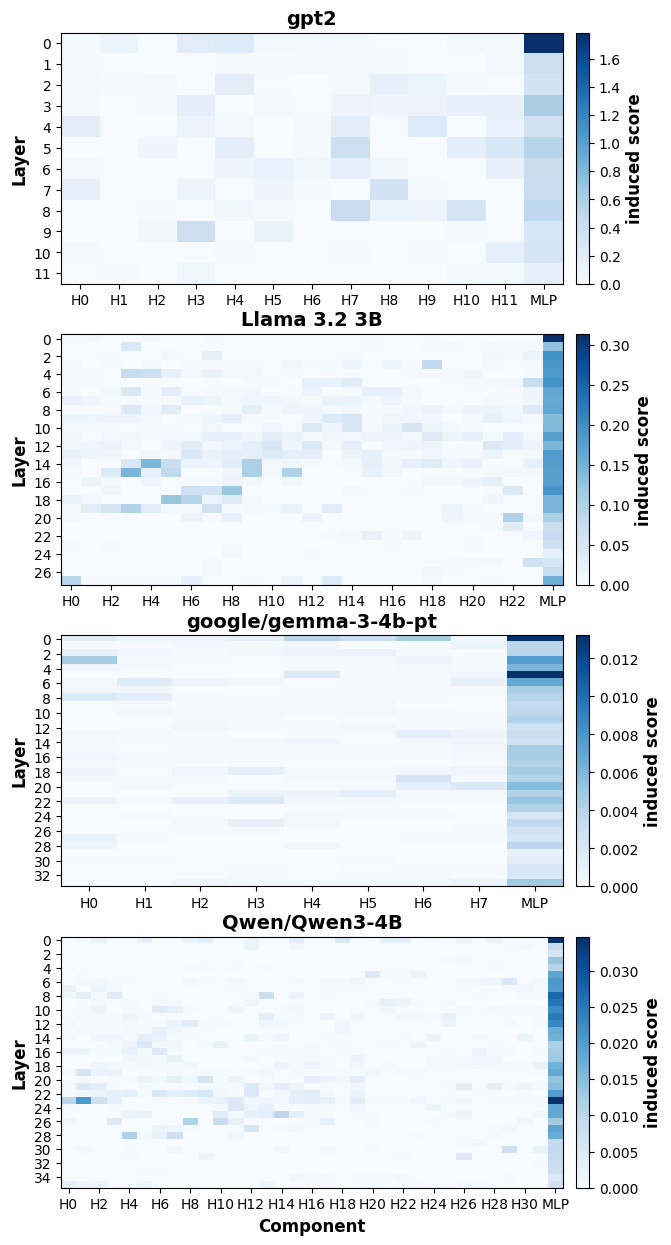}
    \caption{Heatmaps of induced scores for all the components of each model. Best viewed in color.}
    \label{fig:full_circuit_important_components_vertical}
\end{figure}

\subsection{Behavioral Evaluation}
\label{ssec:behavioral_eval}

After selecting the average logit difference as the task metric, we first evaluate each model on the data described in Section~\ref{ssec:dataset_creation} and for each model we retain the samples where it succeeds on the task. A pair is counted as correct if it satisfies Equation~\ref{eq:metric}, because this requires two binary sign conditions to be satisfied simultaneously, the random baseline is \(25\%\). Under this metric, all models performed well above the random baseline: GPT-2 retained 4,721 out of 9,532 pairs, corresponding to 49.53\%; Llama 3.2 3B retained 4,734 pairs (49.66\%); Gemma 3 4B retained 4,371 pairs (45.86\%); and Qwen 3 4B retained 4,300 pairs (45.11\%). These numbers quantify how often each model expressed the intended clean/corrupt animate-inanimate contrast.


Following standard practice, for each model we retain only the samples where the task is successful (see numbers above), and use them to run circuit discovery. By construction, each resulting subset has 100\% accuracy under the selected metric. 
We applied an additional margin threshold (\(>0.5\)) to select high-confidence examples for circuit discovery. For each model-specific dataset, 500 samples were used for discovery and the remainder for evaluation. Detailed sample counts and representative high- and low-margin examples are provided in Appendix~\ref{sec:Dataset_filtering} as well as a circuit discovery run on the shared dataset (Appendix~\ref{app:shared-discovery}).

\subsection{Circuit discovery}

We apply EAP-IG on four open-weight models: Gemma 3 4B, Qwen 3 4B, Llama 3.2 3B, and GPT-2 Small, which we use as a smaller baseline model \citep{gemmateam2025gemma3technicalreport, yang2025qwen3technicalreport, grattafiori2024llama3herdmodels, Radford2019LanguageMA}. We provide further details on EAP-IG in Appendix~\ref{app:eapig-details}.


To rank the edges and later evaluate the resulting circuits, we report both task accuracy and faithfulness. \citet{hanna2024faithfaithfulnessgoingcircuit} showed that tracking accuracy alone is not sufficient to identify a good circuit. For this reason, we follow their experimental design and use faithfulness as the primary metric for both discovery and evaluation. Here, \(m\) denotes the average logit difference defined in Section~\ref{ssec:task-metric}. For each clean-corrupt minimal pair \(x_j\), \(m_{\mathrm{clean}}(x_j)\) is the value of this metric on the clean prefix under the full model, and \(m_{\mathrm{corrupt}}(x_j)\) is the value of the same metric on the corrupt prefix under the full model. \(m_{\mathrm{patched}}(x_j)\) is computed on the corrupt prefix after patching the selected circuit edges from the clean run into the corrupt run. We compute faithfulness separately for each minimal pair:
\[
\mathrm{Faithfulness}(x_j)
=
\frac{
m_{\mathrm{patched}}(x_j) - m_{\mathrm{corrupt}}(x_j)
}{
m_{\mathrm{clean}}(x_j) - m_{\mathrm{corrupt}}(x_j)
}
\]
The faithfulness reported for a circuit is then the average of \(\mathrm{Faithfulness}(x_j)\) over all evaluated minimal pairs.

\begin{figure*}[h]
    \centering
    \includegraphics[width=1\linewidth]{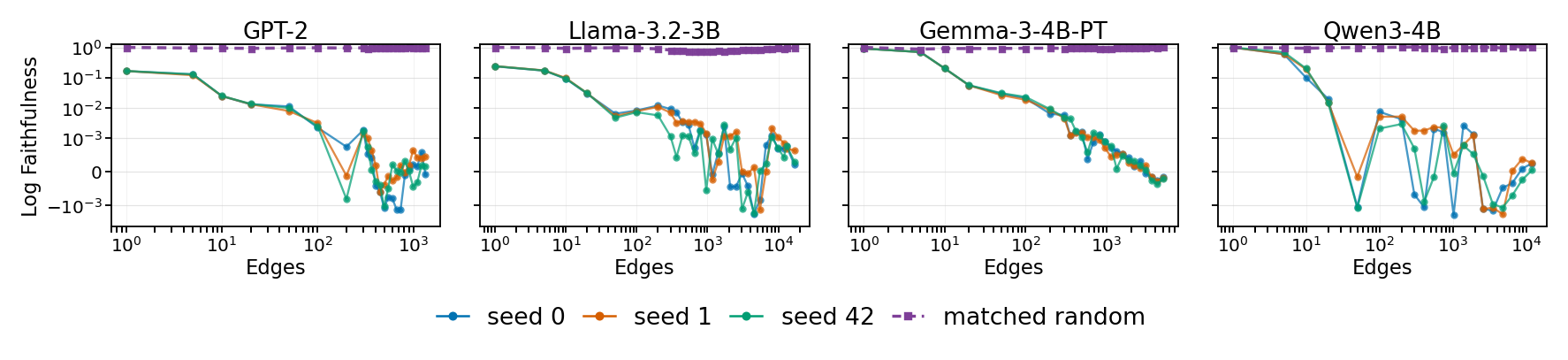}
    \caption{Necessity of ranked EAP-IG circuits compared with matched-random controls. The x-axis displays the amount of ablated edges, ordered by EAP-IG scores. Best viewed in color.}
    \label{fig:necessary_robustness_analysis}
\end{figure*}

For each model, we first ran a full discovery sweep over a range of edge budgets. The sweep used small fixed budgets from 30 to 300 edges, followed by 20 geometrically spaced larger budgets. During circuit selection, following the approach by \citet{hanna2024faithfaithfulnessgoingcircuit}, we used a greedy increasing-budget procedure selecting the smallest sufficient circuit whose faithfulness reaches at least \(85\%\).

\section{Results}

\subsection{Animacy circuit} 

Figure~\ref{fig:circuit_discovery_sweeps} shows that, for each tested model, we do find an animacy circuit, i.e., a set of components that are causally responsible for handling the comprehension of animacy in our task and data. In particular, the lines in the figure show how circuit performance changes as edge budget increases for each model, measured through faithfulness and accuracy. Qwen 3 reaches \(85\%\) faithfulness and \(97\%\) accuracy with only \(1.74\%\) of its edges, while other models require substantially larger circuits, falling in the range of \(7\)--\(13\%\) of total (collapsed) edges.

Zooming into the distribution of induced scores (Figure~\ref{fig:full_circuit_important_components_vertical}), MLPs appear consistently among the most important components across models, particularly the MLP at layer 0. This pattern is especially pronounced in GPT-2, where MLP0 shows a large induced-score gap from the second most important MLP (\(\sim 1.18\)). The gap is smaller in Llama 3.2 (\(\sim 0.11\)); in Gemma, MLP0 and MLP5 have the same induced score of \(0.13\), with a gap of \(\sim 0.06\) from MLP3; and in Qwen 3, MLP0 and MLP23 are the most important (\(\sim 0.034\)), with only a small gap from the third most important MLP (\(\sim 0.007\)). 

Overall, the component-importance plot suggests that the discovered animacy circuit is not concentrated in a single localized region of the model. Instead, the most important components are mostly MLPs, spanning from the first layer toward middle/late layers, while attention heads carry a much smaller share of importance.

\subsection{Is the circuit robust?}
\label{ssec:circuit_robustness}
The budget sweep across the top ranked edges provides us with a sufficient circuit, however we lack guarantees of its effective robustness. For this reason, we complement the budget sweep with additional checks, covering random-edge baselines and stability across multiple discovery runs. Across models, random-edge circuits, i.e., circuits made of randomly sampled edges, do not restore meaningful task performance when retained, nor do they produce comparable degradation when ablated (see Appendix: Figure~\ref{fig:budget_analysis}). At the budget where the discovered circuit first reaches \(85\%\) faithfulness, the faithfulness gap between the discovered circuit and the random baseline is \(87.6\%\), \(93.5\%\), \(87.9\%\), and \(85.6\%\) for GPT-2, Llama 3.2 3B, Gemma 3 4B, and Qwen 3 4B, respectively. To test the reliability of our circuits, we repeat discovery using three different discovery sets obtained from three random seeds \((0,1,42)\). Across all models, the three runs reach the same faithfulness at the same budget and show very high overlap between the circuits found in different runs, with IoU values between \(0.8\)--\(1.0\).

\subsection{Is the circuit necessary?}

Because the discovery procedure identifies circuits that are only sufficient to recover task performance, we also test whether the discovered edges are necessary. 
\textcolor{black}{To do so, we use the same intervention framework adopted for circuit discovery, but test for comprehensiveness rather than sufficiency.} Instead of retaining the top-\(k\) edges and measuring whether they recover model behavior, we ablate cumulative top-\(k\) sets, ordered by EAP-IG scores, from the full model and measure the resulting drop in faithfulness and accuracy. We also perform ablations of size-matched random edge groups, allowing us to test whether the discovered edges are more causally important than random subsets of the graph. 

This analysis shows that only a small number of highly ranked edges are necessary for the task: already at \(k=20\), faithfulness drops below \(0.1\) for all models (see Figure~\ref{fig:necessary_robustness_analysis}), additional ablations show that this effect is specific to the localized top-\(20\) edges, since other disjoint 20-edge sets do not produce comparable damage (Appendix: Table~\ref{tab:conditional-ablation-control}). We also investigate whether a backup circuit exists \citep{mcgrath2023hydraeffectemergentselfrepair, wang2022interpretabilitywildcircuitindirect}, by removing both the core subcircuit and the discovered full circuit from the graph and re-running circuit discovery on the reduced graphs. In both cases (Appendix: Table~\ref{tab:shadow-rediscovery}) we do not find a strong rediscovered backup circuit, further supporting the claim that the identified edges are genuinely necessary for the task in these runs.

\begin{table*}[t]
\centering
\small
\begin{tabular}{lccccccc}
\toprule
& \multicolumn{2}{c}{Task 1: BLiMP + orig targets}
& \multicolumn{3}{c}{Task 2: orig prefixes + named}
& \multicolumn{2}{c}{Task 3: BLiMP + named} \\
\cmidrule(lr){2-3}\cmidrule(lr){4-6}\cmidrule(lr){7-8}
Model & Full LLM & Orig. circuit & Full LLM & Orig. circuit & New circuit & Full LLM & Orig. circuit \\
 & Faith/Acc & Faith/Acc & Faith/Acc &Faith/Acc & Faith/Acc & Faith/Acc & Faith/Acc \\
\midrule
GPT-2 & \underline{1.00}/0.85 & 0.86/\textbf{0.95} & \underline{1.00}/\textbf{1.00} & 0.53/0.69 & 1.03/0.93 & \underline{1.00}/\textbf{0.86} & 0.86/0.05 \\
Gemma 3 4B & \underline{1.00}/\textbf{0.92} & 0.88/0.70 & \underline{1.00}/\textbf{1.00} & 0.79/0.83 & 1.02/0.92 & \underline{1.00}/\textbf{0.84} & 0.88/0.47 \\
Llama 3.2 3B & 1.00/0.89 & \underline{0.92}/\textbf{0.91} & \underline{1.00}/\textbf{1.00} & 0.91/0.89 & 1.54/0.95 & \underline{1.00}/\textbf{0.94} & 0.92/0.17 \\
Qwen 3 4B & \underline{1.00}/\textbf{0.84} & 0.86/0.33 & \underline{1.00}/\textbf{1.00} & 0.60/0.79 & 1.02/0.96 & \underline{1.00}/\textbf{0.94} & 0.86/0.52 \\
\bottomrule
\end{tabular}%
\caption{Circuit transferability and target-task circuit discovery across transfer tasks. "Full LLM" denotes the full model on the target-task, "Orig. circuit" denotes the original \(85\%\)-faithfulness circuit transferred to the target task, and "New circuit" denotes a circuit independently discovered on the target task when possible. Bold marks the highest accuracy within each task while underline denotes the highest faithfulness.}
\label{tab:circuit-transferability}
\end{table*}

\section{Analysis}

\subsection{Functional role of the necessary core components}

\begin{figure}[h]
    \centering
    \includegraphics[width=\linewidth]{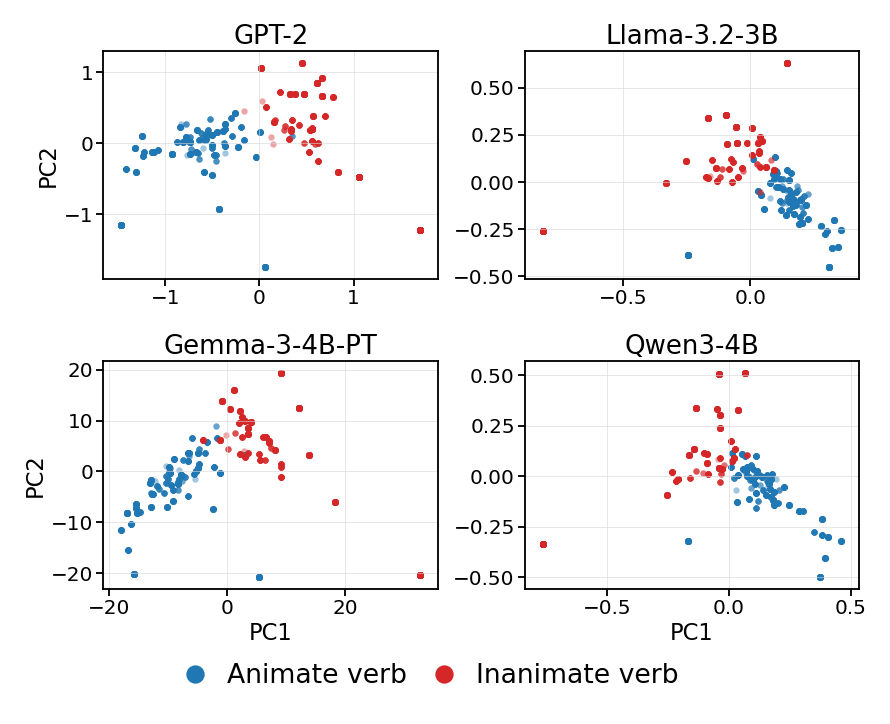}
    \caption{2-dimensional principal components analysis of activations at layer\_0\_pre for the verb token. Best viewed in color.}
    \label{fig:PCA}
\end{figure}

After testing circuit robustness and identifying the necessary subcircuits, we ask what functional role their core components play. From Figures~\ref{fig:fig:circuit1}-\ref{fig:induced_score_necessary} in the Appendix, we observe that MLP0, and especially the input-to-MLP0 connection, receives one of the highest induced scores. We therefore first inspect the verb-token representations at the layer-0 pre-residual stream. The first two principal components show a clear separation between animate and inanimate verbs across models (Figure~\ref{fig:PCA}), suggesting that the relevant verb-level contrast is already available in the earliest residual representation.

However, the logit contribution analysis (Figure~\ref{fig:logit_contribution} in Appendix) shows that MLP0 contributes little directly to the animate-minus-inanimate logit direction, while the strongest direct contributions come mostly from later MLPs. Thus, MLP0 appears to participate in the circuit by routing or preparing information used downstream, rather than by directly writing the animacy decision to the logits. In other words, the verb-level contrast is available early, but becomes logit-aligned only after later transformations.

We next analyze the attention patterns of the most important heads (Figures~\ref{fig:attention_patterns_gpt2}-\ref{fig:attention_patterns_qwen} in Appendix). Across all models, clean and corrupt attention maps are visually very similar, suggesting that the animacy-relevant difference is carried mainly by the representations read from attended positions, not by changes in where heads attend. In general, the core of common edges shares a similar pattern across models. Animacy-relevant information is available early, especially around verb representations and the MLP0 pathway, but early components appear to prepare or route verb-related information rather than directly writing the final animate--inanimate preference to the logits. Later MLPs are more directly aligned with the output decision, while attention heads mostly support positional routing, such as possible verb-to-\textit{by}-to-\textit{the} pathways, rather than encoding animacy through changes in attention weights alone.

\subsection{A task-specific or a general-purpose animacy circuit?}

\citet{mueller2024missedcausesambiguouseffects} and \citet{zhang2024bestpracticesactivationpatching} emphasize an important limitation of circuit-discovery methods: the circuits they recover can depend strongly on the dataset used to define the task. For this reason, we test whether the circuits discovered on our original animacy-completion task transfer beyond that exact discovery setting. Concretely, we test our circuits on: (1) \textit{different data} (\texttt{subject\_passive} prefixes from BLiMP), but \textit{same expected completion}, i.e., our animate and inanimate targets; (2) \textit{same data}, but \textit{different expected completion}, i.e., animate and inanimate named entities;~\footnote{To do so, we remove the final determiner from the passive prefixes in our data, from ``\textit{The [patient] was [verb] by the}'' to ``\textit{The [patient] was [verb] by}'', making them compatible with a named entity continuation.} (3) \textit{different data} (\texttt{subject\_passive}  from BLiMP), and \textit{different expected completion}, i.e., animate/inanimate named entities. We refer to these generalizations as Setup 1, 2, and 3, respectively.



In all these settings, we evaluate the original \(85\%\)-faithfulness circuits as well as the full models, so that circuit failures can be distinguished from model failures. Table~\ref{tab:circuit-transferability} shows how in setup 1, all the circuits generalize extremely well except for Qwen (which reaches only \(33\%\) of accuracy), showing their capability to handle animacy in different data. In setup 2, the circuits generalize well across all models. Suggesting that the recovered circuits can still support animacy completion when the target sets are made of named entities. However, while accuracy remains relatively high, faithfulness is substantially lower than in the original setup. Finally, in setup 3, we notice that, across models, the circuits do not generalize. This shows that, if both the data and expected completion change, then the circuit breaks and is no longer capable of picking the right (animate or inanimate) continuation.

Overall, these results indicate that the discovered circuits are not strictly tied to the original setup: in some settings, especially when only the data or only the target type changes, they transfer well and preserve part of the animacy behavior. At the same time though, this transfer is not uniform. When both the prefix distribution and the expected completions change, the circuits often fail even when the full models are still able to perform the task. This means that recovered circuits capture part of the animacy computation, while other aspects of animacy are handled by other parts of the models.

\textcolor{black}{\paragraph{Another animacy circuit.}
Since setup 2 preserves the original minimal-pair structure, it is also suitable for circuit discovery. We therefore run discovery directly on this setup and compare the resulting circuit with the original common-noun animacy circuit. As expected (Table~\ref{tab:circuit-transferability}), these newly discovered circuits achieve higher accuracy. However, compared with the original circuits, they still share a substantial portion of edges, with an IoU of \(\sim0.4\)--\(0.7\) (see Appendix Table~\ref{tab:task2-circuit-overlap}).
}

\section{Discussion}

In this work, we investigated whether animacy completion in large language models is supported by a localized causal circuit, and whether this circuit generalizes beyond the data used to find it. Using a controlled dataset designed to isolate animacy-sensitive behavior, we applied EAP-IG to four language models. Across models, we found circuits that recover the source behavior and contain smaller necessary cores whose ablation is strongly damaging. This shows that animacy-sensitive completion is a behaviour that can be causally localized.


At the same time, the recovered circuits should not be interpreted as a compact and universal "animacy mechanism". These circuits are larger and more distributed than circuits found for other tasks such as Greater-Than, IoI, and Country-Capital \citep{greater-than, wang2022interpretabilitywildcircuitindirect, hanna2024faithfaithfulnessgoingcircuit}. This is likely due to the multifactorial nature of animacy: animacy completion requires combining lexical semantics, event plausibility, and expectations about likely agents or causes. The necessary cores suggest a common pattern across models, with verb-related information available early around MLP0, later MLPs more aligned with the final animate-inanimate preference, and attention heads mostly supporting positional routing.



The transfer experiments further qualify this result. The circuits are stable across discovery samples and outperform random edge sets, but their generalization is uneven and model-dependent. This may be especially relevant for Qwen, whose recovered circuit is much smaller than the others and may therefore be more sensitive to changes in the data distribution. They transfer when only the data distribution or target type changes, but often fail when both the prefix distribution and expected completions change, even when the full models still solve the task. This suggests that they capture a robust part of the animacy computation, but one that remains tied to the task, data, and intervention setup.

In conclusion, we found a powerful and robust circuit for animacy-sensitive behavior, but not \textit{the} circuit for animacy. Animacy may be too multifaceted, graded, and context-dependent to be captured by a single circuit that generalizes uniformly across settings. Future work is needed to shed further light on this problem.

\section*{Limitations}

Our study isolates a specific form of animacy-sensitive completion: passive prefixes in which the verb shifts the expected continuation toward either an animate agent or an inanimate cause. Moreover, our animate target set is restricted to person nouns, so the operational contrast is between plausible human agents and inanimate causes, not  between all levels of the animacy hierarchy. This controlled setting does not establish that the same computation supports animal or personified agents, coreference, agreement, thematic-role assignment in active sentences, or discourse-level interpretation. Future work should test whether the recovered structure extends to these phenomena and to more graded notions of animacy.

The dataset also depends on generated semantic frames and automatic plausibility filtering. The frames were generated by a checkpoint not included among the evaluated models and were manually inspected, reducing direct circularity and obvious semantic failures. These precautions do not eliminate lexical, event-level, or generator-specific regularities. Stronger tests would use larger human-validated datasets, lexically or frame-disjoint evaluation splits, better control of frequency and plausibility, and sensitivity analyses over filtering thresholds.

Our circuit results are also conditional on successful model behavior. Circuit discovery uses 500 sign-correct, high-confidence pairs with margin \(>0.5\), whereas evaluation (Section~\ref{ssec:behavioral_eval}) uses all remaining sign-correct pairs, including low-margin cases. Pairs on which the model fails either sign condition are excluded. Thus, the evaluation tests whether a circuit discovered on high-confidence behavior generalizes to weaker instances of the same successful behavior, but it does not explain model failures or estimate the mechanism's prevalence over the full dataset.

The recovered circuits should not be interpreted as complete or unique descriptions of the models' computation. EAP-IG provides an approximate ranking of candidate edges, which we subsequently evaluate through explicit patching and ablation. Nevertheless, the results depend on the task metric, clean--corrupt contrast, graph representation, and intervention distribution. We therefore interpret the circuits as task-relative causal subgraphs: they are sufficient to recover much of the behavior under our intervention setup and contain a compact core whose ablation is strongly damaging, but other task formulations or intervention schemes may identify different structures.

Finally, the functional interpretation of this necessary core remains preliminary. Our analyses suggest that verb-related distinctions are available in early representations and become more directly aligned with the animate--inanimate logit preference through later components, but they do not identify the represented features or fully characterize their transformation across layers.

\section*{Acknowledgments}
We are grateful to the members of the Multimodality, Language, and Interpretability (Mulini) lab at the University of Amsterdam for the feedback on the work, particularly to Michael Hanna for his suggestions about the dataset construction and metrics for the task. This publication was made possible through funding from the Sector Plan for Medicine and Health Sciences, supported by the Dutch Ministry of Education, Culture and Science (OCW).

\clearpage

\bibliography{custom}

\clearpage

\appendix

\section*{Appendix Roadmap}

The supplementary material is organized as follows:
\begin{itemize}
    \item \textbf{Appendices A--D:} Dataset construction, target construction,
    metric selection, and dataset validation and filtering.
    
    \item \textbf{Appendix E:} EAP-IG methodology, computational-graph
    representation, edge grouping, and component scoring.
    
    \item \textbf{Appendices F--G:} Circuit necessity, random-edge controls,
    backup-circuit analysis, and stability across discovery runs.
    
    \item \textbf{Appendices H--I:} Transfer-dataset construction, transfer
    evaluation, and target-task circuit rediscovery.
    
    \item \textbf{Appendices J--M:} Necessary-core visualization and
    exploratory component-level characterization.
    
    \item \textbf{Appendix N:} Circuit discovery on model-specific versus
    shared retained datasets.
\end{itemize}

\section{Dataset Construction}
\label{sec:dataset_building}

\subsection{GPT 5.4 prompt}
\textbf{Prompt:}
\begin{quote}
    "\textit{
    You are an expert computational linguist. I am building a psycholinguistics dataset to test animacy-based s-selection in language models. I need you to generate 10 specific "Intersection Frames" of vocabulary. In an Intersection Frame, the "Patient" noun must logically be able to be acted upon by BOTH a "Clean Verb" (which implies a human intentional agent) AND a "Corrupt Verb" (which implies a physical force, natural event, or inanimate agent). Both resulting prefixes MUST be perfectly plausible and natural in English. 
    \\\\
    The sentence template is: "The [PATIENT] was [VERB] by the "
    \\\\
    FEW-SHOT EXAMPLES OF DESIRED LOGIC:
    \\\\
    Frame: Humans as Physical Entities \\
    Patients: victim, pedestrian, driver, citizen, survivor, boy, girl, teacher \\
    Clean Verbs (Human Agent): rescued, treated, questioned, arrested \\
    Corrupt Verbs (Force Agent): struck, crushed, trapped, blinded \\
    Result: \\
    Clean: "The victim was rescued by the " (Plausible) \\
    Corrupt: "The victim was crushed by the " (Plausible)
    \\\\
    Frame: Physical Structures \\
    Patients: building, bridge, house, vehicle, vessel, rails, train, runway \\
    Clean Verbs (Human Agent): inspected, purchased, designed, defended \\
    Corrupt Verbs (Force Agent): destroyed, shattered, flooded, demolished \\
    Result: \\
    Clean: "The bridge was designed by the " (Plausible) \\
    Corrupt: "The bridge was destroyed by the " (Plausible) \\
    \\\\
    Frame: Information \& Resources \\
    Patients: record, file, document, system, network \\
    Clean Verbs (Human Agent): reviewed, audited, published, translated \\
    Corrupt Verbs (Force Agent): deleted, corrupted, erased, overloaded \\
    Result: \\
    Clean: "The file was audited by the " (Plausible) \\
    Corrupt: "The file was corrupted by the " (Plausible)\\
    \\\\
    YOUR TASK: \\
    Generate exactly 10 distinct Intersection Frames (you may expand on the 3 examples above and invent more, such as a "Biological/Nature" frame or a "Landscape/Terrain" frame). For EACH of the 10 frames, generate exactly: \\
    20 Patients (Nouns) \\
    10 Clean Verbs \\
    10 Corrupt Verbs
    \\\\
    STRICT CONSTRAINTS: \\
    All words must be single English words. No multi-word phrases.
    All verbs MUST be in the past-participle form (e.g., "rescued", not "rescue").
    Output ONLY valid JSON in this exact format:\\
    \{"frames": [\{"name": "Frame Name", "patients": ["...", "..."], "clean\_verbs": ["...", "..."], "corrupt\_verbs": ["...", "..."]\}, ...]\} \\
    Do not include any markdown formatting, code blocks, or introductory/concluding text. Just the raw JSON object.
}"
\end{quote}

We defined as semantic frames the following categories: Humans as Physical Entities, Physical Structures, Information Resources, Biological Organisms, Vehicles and Conveyances, Financial Economic, Academic Scientific, Culinary, Legal Administrative, and Textile Garment. Each frame contains: patients (i.e., passive subjects in that domain), clean\_verbs (i.e., past participles favoring animate agents), and corrupt\_verbs (i.e., past participles favoring inanimate causes). We manually evaluated each frame, checking all the verbs and patients contained in them to verify their correctness and consistency.

\paragraph{LLM-generated semantic frames.}
The fact that initial semantic frames used to construct the dataset were generated with an LLM introduces a possible source of bias, since the generated patients and verbs may reflect the generator model's own lexical or animacy-related preferences. To reduce this risk, we used the generator only to propose candidate lexical material; the final task labels are determined by the clean--corrupt sentence construction and by independently constructed WordNet-based animate and inanimate target sets. We also manually inspected the generated frames and applied a semantic-plausibility filter to both clean and corrupt prefixes, retaining only pairs for which both prefixes were plausible. Importantly, the model used to generate the semantic frames (GPT 5.4) was not one of the models evaluated in our circuit-discovery experiments. The evaluated models were GPT-2 Small, Llama 3.2 3B, Gemma 3 4B, and Qwen 3 4B. This reduces the risk that the discovered circuits simply reflect idiosyncrasies of the generator model, although it does not fully eliminate possible dataset-generation biases.

\subsection{Semantic filtering prompt}
\textbf{Prompt:}
\begin{quote}
    "\textit{
    Does the following incomplete sentence prefix make logical and semantic sense so far? \\
    Ignore the fact that it ends abruptly. \\
    Answer strictly with 'Yes' or 'No'. \\
    Prefix: "\{prefix\}"\\
    Answer:
    }"
\end{quote}

considering the following token variants:
\begin{itemize}
    \item Yes variants: " Yes", "Yes", "yes", " yes"
    \item No variants:  " No", "No", "no", " no"
\end{itemize}

For each clean and corrupt prefix, we compute the probability assigned to the \texttt{Yes} token and use it as a plausibility score.

\subsection{Semantic score distribution}
\label{ssec:scoring}

\begin{figure}[h]
    \centering
    \includegraphics[width=1\linewidth]{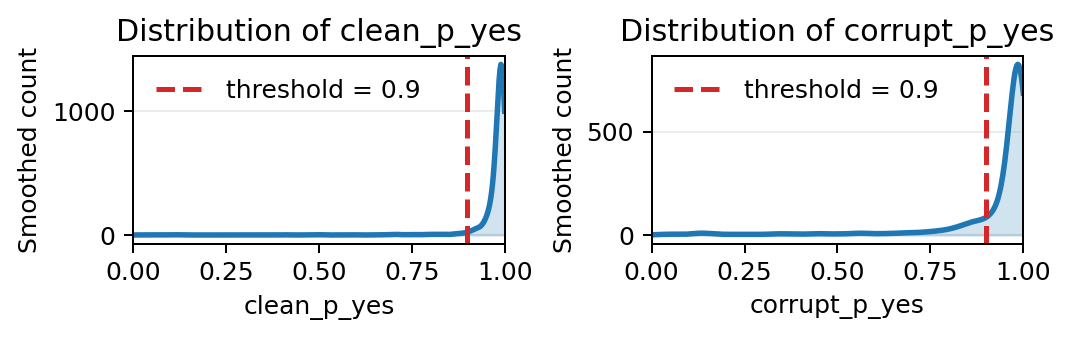}
    \caption{Distribution of semantic scores for animate and inanimate sentence groups.}
    \label{fig:prob_score_curves}
\end{figure}

As shown in Figure~\ref{fig:prob_score_curves}, the semantic scores for both sentence groups are strongly concentrated near 1.0, with only a short left tail toward lower scores. This indicates that the vast majority of sentences in both the animate and inanimate conditions received high semantic plausibility scores, for this reason we decided to retain a minimal pair only if both its clean and corrupt prefixes receive a plausibility score above 0.90.

To further assess the reliability of the semantic scoring procedure, we manually inspected 200 minimal pairs sampled from the pre-filtered dataset. All inspected pairs were judged to be semantically meaningful, and the model assigned comparable scores to the two members of each pair.

\begin{table*}[!ht]
\centering
\small
\begin{tabular}{lrrrr}
\toprule
Model & Retained examples & Margin $>0.5$ candidates & Discovery & Evaluation \\
\midrule
GPT-2 & 4,721 & 4,700 & 500 & 4,221 \\
Llama 3.2 3B & 4,734 & 4,703 & 500 & 4,234 \\
Gemma 3 4B & 4,371 & 4,339 & 500 & 3,871 \\
Qwen 3 4B & 4,300 & 4,254 & 500 & 3,800 \\
\bottomrule
\end{tabular}
\caption{Model-specific dataset sizes after sign-based filtering under the average logit difference metric. The margin threshold is used only to define the
high-confidence pool from which 500 discovery examples are sampled; all remaining retained examples are used for evaluation.}
\label{tab:model_specific_dataset_sizes}
\end{table*}

\section{Target construction}
\label{app:target-construction}

To build the animate and inanimate target sets we followed a procedure similar to the one presented by \citet{gregorio-etal-2025-cross}. An animate candidate is retained only if all of its noun senses have the WordNet lexname \texttt{noun.person}, at least one hypernym path reaches \texttt{person.n.01}, and no sense has \texttt{group.n.01} in its hypernym closure. While we defined inanimate targets as non-person nouns: a candidate is kept only if none of its noun senses is \texttt{noun.person}, all noun senses belong to the allowed inanimate lexnames \texttt{noun.event}, \texttt{noun.phenomenon}, \texttt{noun.state}, \texttt{noun.artifact}, \texttt{noun.substance}, or \texttt{noun.object}, and no sense has \texttt{group.n.01} in its hypernym closure. Candidate words are then further normalized by lowercasing them, keeping only single alphabetic words with no spaces or hyphens, singularizing them with \texttt{inflect}, and requiring them to be single-token under the construction tokenizer (initially GPT-2).

\begin{table*}[b]
\centering
\small
\begin{tabular}{llrrrr}
\toprule
Group & Minimal pair & GPT-2 & Qwen & Llama & Gemma \\
\midrule
High margin &
\begin{tabular}[c]{@{}l@{}}
The robe was tailored by the \\
The robe was faded by the
\end{tabular}
& 3.422 & 4.222 & 4.121 & 4.452 \\

High margin &
\begin{tabular}[c]{@{}l@{}}
The tree was planted by the \\
The tree was choked by the
\end{tabular}
& 3.522 & 3.464 & 3.268 & 5.237 \\

High margin &
\begin{tabular}[c]{@{}l@{}}
The house was planned by the \\
The house was battered by the
\end{tabular}
& 3.741 & 3.849 & 3.605 & 5.054 \\

\midrule

Low positive margin &
\begin{tabular}[c]{@{}l@{}}
The program was reviewed by the \\
The program was encrypted by the
\end{tabular}
& 0.093 & 0.053 & 0.097 & 0.082 \\

Low positive margin &
\begin{tabular}[c]{@{}l@{}}
The appeal was reviewed by the \\
The appeal was blocked by the
\end{tabular}
& 0.135 & 0.168 & 0.194 & 0.095 \\

Low positive margin &
\begin{tabular}[c]{@{}l@{}}
The allowance was protected by the \\
The allowance was disrupted by the
\end{tabular}
& 0.194 & 0.218 & 0.246 & 0.150 \\
\bottomrule
\end{tabular}
\caption{Examples of minimal pairs from the dataset with high AVG-LD margins across all models and low positive AVG-LD margins across all models. Margins are computed as clean AVG-LD minus corrupt AVG-LD.}
\label{tab:minimal-pair-examples}
\end{table*}

\section{Metric Investigation}
\label{sec:metric_investigation}
\begin{figure*}[t]
    \centering
    \includegraphics[width=1\linewidth]{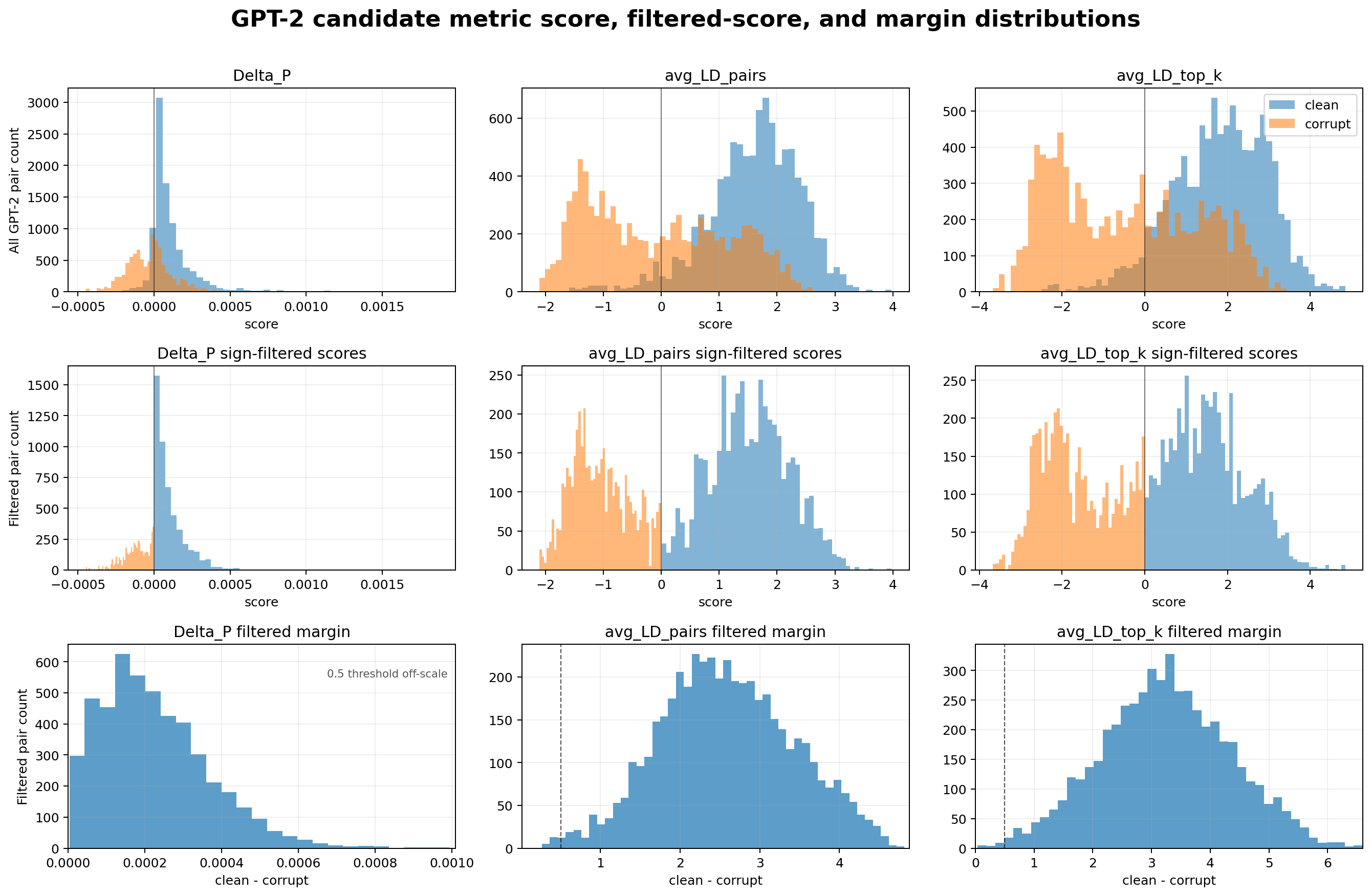}
    \caption{GPT-2 score distributions for the candidate task metrics. Columns show \(\Delta Probability\), pairwise average logit difference, and top-\(k\) average logit difference. The first row reports clean and corrupt score distributions over all candidate pairs; the second row shows the same distributions after success-metric filtering; and the third row shows the resulting clean--corrupt margin distributions. The vertical reference lines indicate the decision boundary and margin threshold used to assess whether a metric separates the two conditions.}
    \label{fig:gpt2_metrics}
\end{figure*}

\begin{figure*}[b]
    \centering
    \includegraphics[width=1\linewidth]{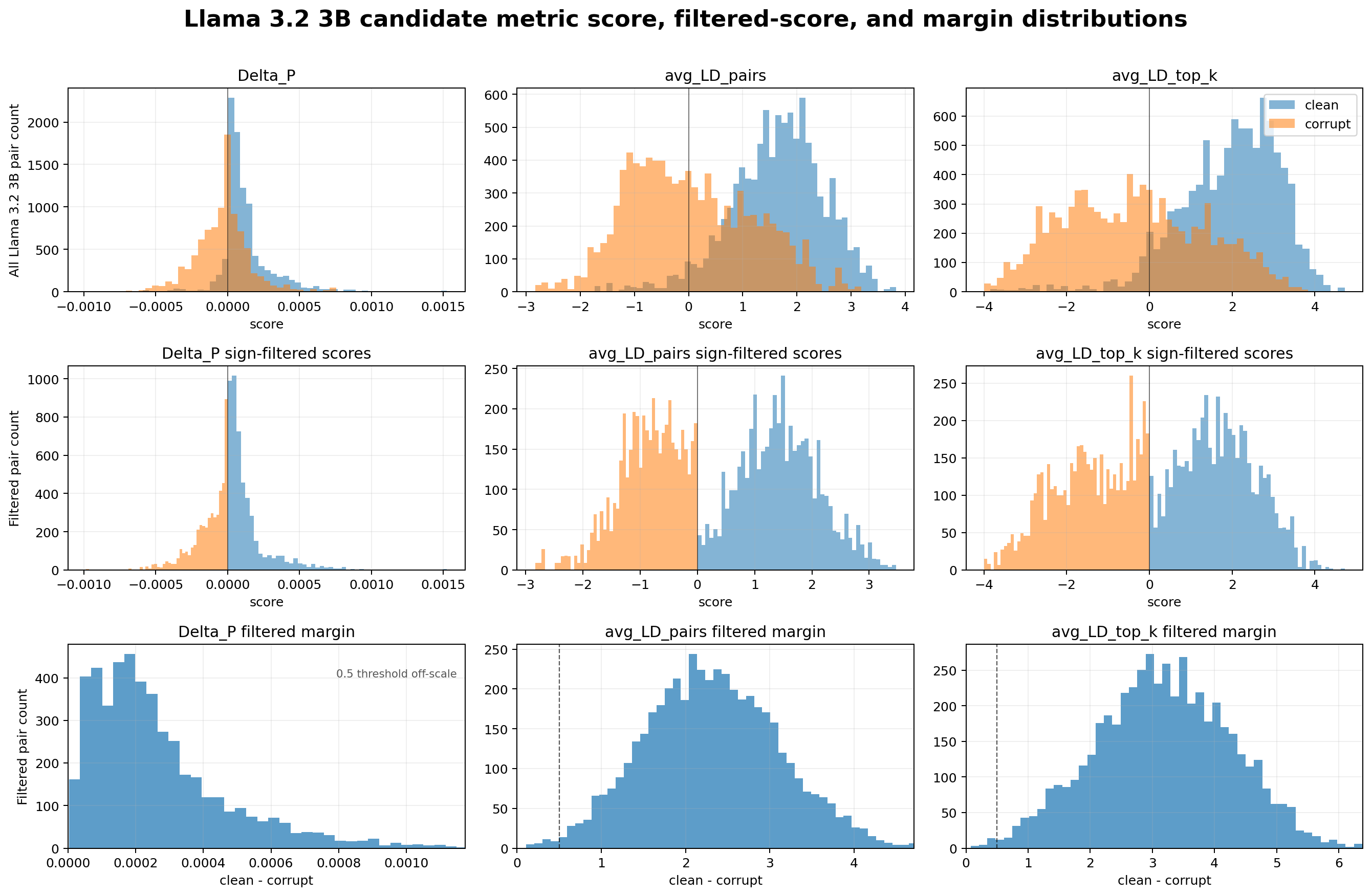}
        \caption{Llama 3.2 score distributions for the candidate task metrics.}
    \label{fig:llama_metrics}
\end{figure*}

\begin{figure*}
    \centering
    \includegraphics[width=1\linewidth]{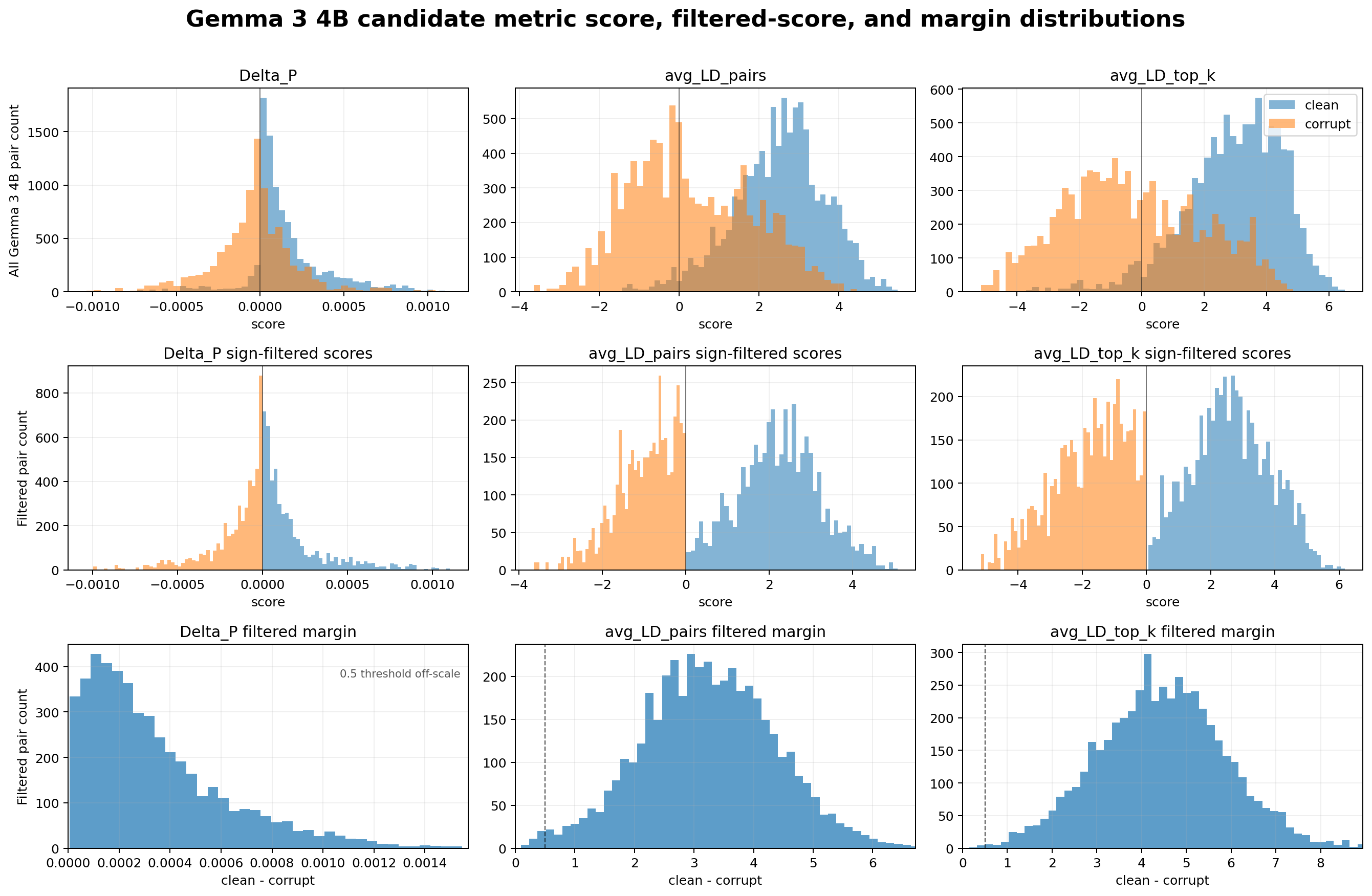}
        \caption{Gemma 3 score distributions for the candidate task metrics. Columns show \(\Delta Probability\), pairwise average logit difference, and top-\(k\) average logit difference. The first row reports clean and corrupt score distributions over all candidate pairs; the second row shows the same distributions after success-metric filtering; and the third row shows the resulting clean--corrupt margin distributions. The vertical reference lines indicate the decision boundary and margin threshold used to assess whether a metric separates the two conditions.}
    \label{fig:gemma_metrics}
\end{figure*}

\begin{figure*}
    \centering
    \includegraphics[width=1\linewidth]{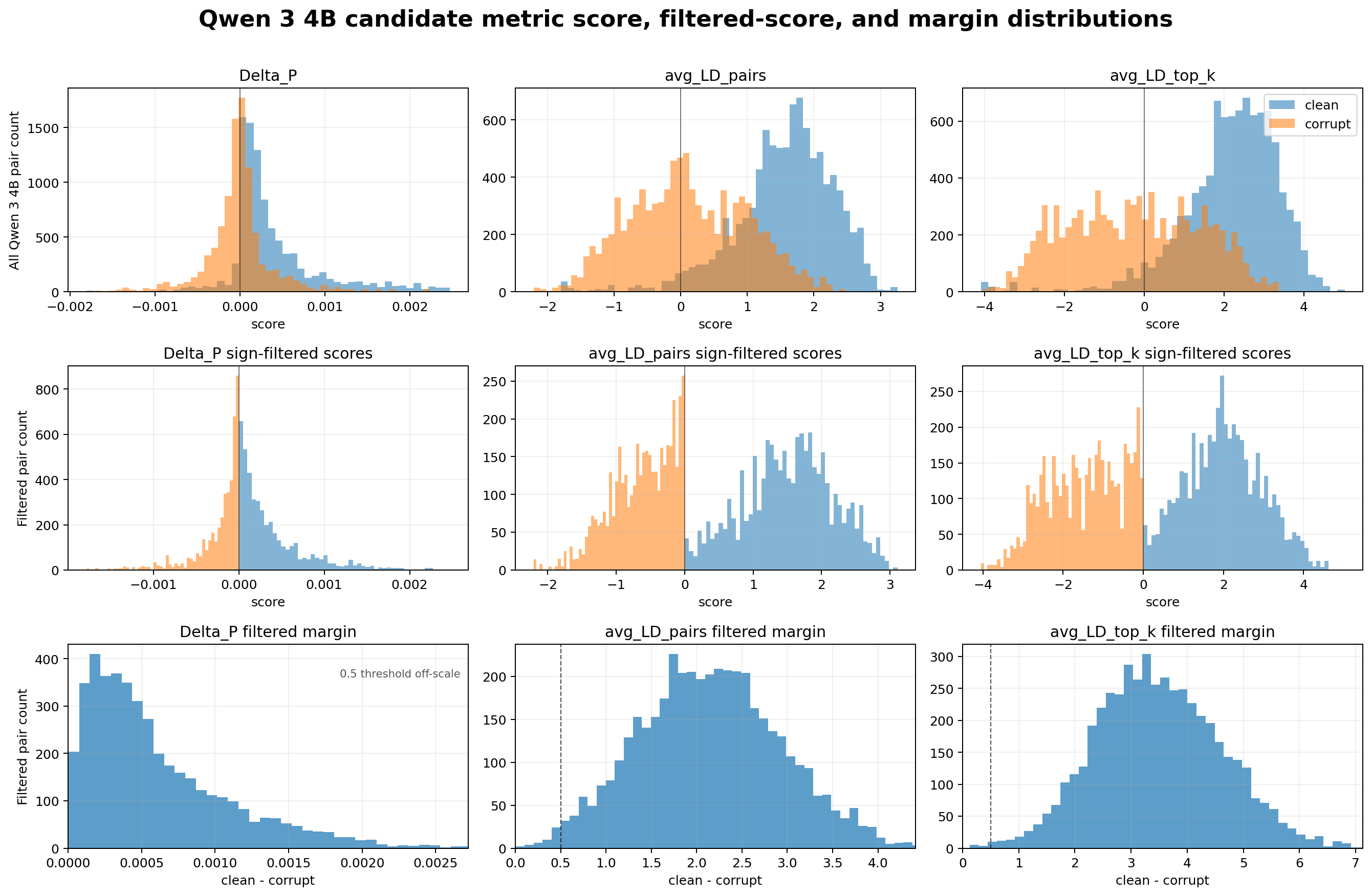}
        \caption{Qwen 3 score distributions for the candidate task metrics.}
    \label{fig:qwen_metrics}
\end{figure*}

Following good practices in mechanistic interpretability, before selecting the average logit difference as our main task metric, we evaluated several alternative set-based metrics. In addition to the full average logit difference, we tested a top-\(k\) version of the same quantity, computed by averaging only over the highest-scoring animate and inanimate targets within each set, and an average probability difference, defined as the mean probability assigned to animate targets minus the mean probability assigned to inanimate targets. The top-\(k\) metric was motivated by the possibility that averaging over the full target sets might dilute the signal through many contextually irrelevant low-logit words, whereas the probability-based metric offered a more direct interpretation in probability space. For each metric, we computed separate scores for clean and corrupt prefixes and evaluated both the sign-based success criterion and the resulting margin distributions across models. Based on these comparisons, we selected the full average logit difference because it directly captures the animate--inanimate contrast over the complete target sets, remains stable across models, and avoids the numerical compression observed for probability-based metrics. Detailed plots and comparisons are provided in Figures \ref{fig:gpt2_metrics}-\ref{fig:qwen_metrics}.

\section{Dataset success filtering}
\label{sec:Dataset_filtering}

Table~\ref{tab:model_specific_dataset_sizes} summarizes the model-specific dataset sizes after success filtering and the resulting discovery and evaluation splits. Table~\ref{tab:minimal-pair-examples} presents representative minimal pairs with high and low positive AVG-LD margins.


\newpage
\onecolumn

\clearpage
\twocolumn
\section{EAP-IG}
\label{app:eapig-details}

For circuit discovery, we use the computational graph representation adopted in EAP-IG \citep{hanna2024faithfaithfulnessgoingcircuit}. A model is represented as a directed acyclic graph \(G=(V,E)\), where nodes \(V\) correspond to model components, such as attention heads, MLPs, the input node, and the logits node, and edges \(E\) represent directed information flow between components. For a node \(v\), its input is the sum of the outputs of all parent nodes connected to it:
\[
x_v = \sum_{(u,v)\in E} z_u .
\]

EAP estimates the importance of an edge \(e=(u,v)\) by approximating the effect of replacing the clean activation on that edge with the corresponding corrupted activation. Let \(z_u\) be the output of node \(u\) on the clean input, and \(z'_u\) the output of the same node on the corrupted input. Given a task metric \(M\), we define the loss as \(L=-M\), so that lower loss corresponds to better task performance. Standard EAP assigns the edge the first-order score
\[
\mathrm{EAP}(e)
=
(z'_u-z_u)^\top \nabla_{x_v} L ,
\]
where \(\nabla_{x_v} L\) is the gradient of the loss with respect to the input of node \(v\). This gives a scalable approximation to activation patching, since all edge scores can be estimated with a small number of forward and backward passes.

EAP-IG improves this approximation by replacing the single local gradient with an average of gradients along the interpolation path between corrupted and clean activations. For \(m\) interpolation steps, the score is
\begin{gather}
\mathrm{EAP\mbox{-}IG}(e) 
= \notag \\
(z'_u-z_u)^\top
\frac{1}{m}
\sum_{k=1}^{m}
\nabla_{x_v} L\!\left(
z'_u + \frac{k}{m}(z_u-z'_u)
\right). \notag
\end{gather}
In practice, the interpolation is applied to the relevant edge activations within the full forward pass, following the implementation of \citet{hanna2024faithfaithfulnessgoingcircuit}. Intuitively, this reduces the risk that an edge receives a low score only because the gradient at the clean activation is locally small, even though changing the edge activation would affect the task metric elsewhere along the clean--corrupt path.

In our experiments, EAP-IG produces scores for the low-level edges of the computational graph. However, several low-level edges can correspond to the same visible connection between model components. We therefore collapse these low-level edges into component-level edge groups before ranking. Each collapsed edge group \(g\) represents a directed connection between two visible graph nodes, such as an attention head, MLP layer, input node, or logits node. We assign each group a score equal to the total absolute attribution mass of its underlying edges:
\[
S(g) = \sum_{e\in g} |\mathrm{EAP\mbox{-}IG}(e)| .
\]
All edge budgets reported in the paper therefore refer to collapsed edge groups rather than individual low-level graph edges.

Finally, to summarize which components participate most strongly in the ranked edge set, we define an induced score for each node. Let \(\mathcal{N}(v)\) be the set of collapsed edge groups incident to node \(v\), that is, the collapsed edges for which \(v\) is either the parent or the child node. The induced score of \(v\) is then
\[
I(v) = \sum_{g \in \mathcal{N}(v)} S(g).
\]
Thus, a component receives a high induced score when it is connected to high-attribution collapsed edges, either as their source or as their target. We use induced scores only as a descriptive summary of component importance; circuit discovery and budget selection are performed over collapsed edge groups.

\section{Circuit sufficiency and comprehensiveness analysis}
\label{sec:circuit_analysis}

\subsection{Circuit necessity analysis}
\label{ssec:circuit_necessity_analysis}

To test whether the discovered circuits contain edges that are necessary for the task, we run a cumulative top-\(k\) ablation analysis. We use the same clean--corrupt intervention framework used for circuit discovery, but evaluate comprehensiveness rather than sufficiency. Instead of retaining only the top-\(k\) edges and measuring whether they recover model behavior, we ablate selected edges from the full model by replacing their clean activations with the corresponding corrupt activations, and measure the resulting drop in faithfulness and accuracy.

For each model, edges are ordered by their EAP-IG importance scores. We then progressively ablate the top-\(k\) edges for increasing values of \(k\). Across GPT-2, Llama 3.2 3B, Gemma 3 4B, and Qwen 3 4B, ablating only a small number of highly ranked edges produces a large degradation in performance. In particular, at \(k=20\), faithfulness drops below \(0.1\) for all models. This suggests that the discovered circuits contain a compact necessary core: although the sufficient circuits are often large and distributed, a much smaller set of highly ranked edges is crucial for preserving the behavior.

\subsection{Conditional ablation control}
\label{sec:conditional_ablation_control}

As shown by Table~\ref{tab:conditional-ablation-control} a large drop from ablating the top-\(20\) edges alone does not rule out the possibility that many other similarly sized sets of important edges would be equally damaging. To test this, we compare the localized top-\(20\) edges against other disjoint 20-edge sets drawn from the remaining high-ranked edges.

For each model, we first construct sliding-window rank bands over the EAP-IG ordering, using ranks \(21\)--\(40\), \(41\)--\(60\), and so on until rank 200. We then also sample 100 disjoint 20-edge sets from the remaining ranked edges. Each set is ablated using the same intervention procedure as in the top-\(k\) necessity analysis, and we measure the resulting faithfulness and accuracy.

Across all four models, no alternative disjoint 20-edge set matches the damage caused by ablating the localized top-\(20\) circuit. For example, in Qwen 3 4B, ablating the top-\(20\) edges reduces faithfulness to \(0.015\), while the most damaging sliding-window and sampled alternatives reach only \(0.583\) and \(0.749\), respectively. This shows that the top-\(20\) edges are not merely one arbitrary important 20-edge set among many, but form a specifically damaging core subcircuit for animacy completion behavior.

\section{Discovery of Backup Circuits After Original-Circuit Ablation}
\label{app:shadow-rediscovery}

As an additional check on circuit necessity, we tested whether a backup circuit could be rediscovered after removing the identified circuit from the graph \citep{mcgrath2023hydraeffectemergentselfrepair, wang2022interpretabilitywildcircuitindirect}. For each model, we started from the saved full-model EAP-IG run on the same model-specific dataset split used in the main experiments. We reconstructed the same discovery and validation split, removed the collapsed edges belonging to the selected source circuit, and then reran EAP-IG on the remaining graph.

The removed source circuit was defined as the smallest circuit budget whose saved full-model sweep reached $\texttt{faithfulness\_mean} \geq 0.85$. This is the same criterion used for selecting the sufficient circuits in the main experiments. We also repeat the same rediscovery diagnostic after removing only the localized top-\(20\) necessary subcircuit. In both cases, we rank the surviving collapsed edges and evaluate the best rediscovered circuits with the same validation budget sweep.

Table~\ref{tab:shadow-rediscovery} shows that rediscovery remains near zero across all four models after removing the full selected source circuit. The maximum reduced-graph faithfulness is $0.000433$ for GPT-2, $0.000183$ for Llama 3.2 3B, $0.000066$ for Gemma 3 4B, and $0.000029$ for Qwen 3 4B. In all cases, none of the removed collapsed edges reappeared in the rediscovered ranking. Thus, under the same discovery procedure and evaluation metric used in the main experiments, removing the selected source circuit does not reveal a strong compact backup circuit. The same conclusion holds when removing the localized top-\(20\) necessary subcircuit.

This diagnostic does not rule out every possible form of distributed redundancy: a backup mechanism could require a different objective, a different dataset distribution, or a much larger non-compact set of edges. However, within the EAP-IG setup used here, the remaining graph does not support a rediscovered circuit with meaningful faithfulness. This supports the interpretation that the identified circuit is not merely one arbitrary high-scoring edge set among many equivalent alternatives.



\begin{table*}[t]
\centering
\small
\begin{tabular}{lrrrrrrr}
\toprule
& & \multicolumn{3}{c}{Faithfulness after ablation} &
\multicolumn{3}{c}{Accuracy after ablation} \\
\cmidrule(lr){3-5}\cmidrule(lr){6-8}
Model & $n_{\mathrm{val}}$
& Top-20 & Best rank band & Best sample
& Top-20 & Best rank band & Best sample \\
\midrule
GPT-2
& 4{,}221
& 0.014 & 0.712 & 0.758
& 0.019 & 0.871 & 0.932 \\
Llama 3.2 3B
& 4{,}234
& 0.031 & 0.741 & 0.760
& 0.046 & 0.912 & 0.933 \\
Gemma 3 4B
& 3{,}871
& 0.056 & 0.696 & 0.764
& 0.110 & 0.904 & 0.945 \\
Qwen 3 4B
& 3{,}800
& 0.015 & 0.583 & 0.749
& 0.038 & 0.805 & 0.965 \\
\bottomrule
\end{tabular}
\caption{
Conditional ablation control for top-20 circuit necessity. For each model, we ablate either the localized top-20 EAP-IG edges, the most damaging disjoint adjacent 20-edge rank band from ranks 21--200, or the most damaging of 100 sampled disjoint 20-edge sets from the same candidate range. Lower faithfulness and accuracy indicate a more damaging ablation. No rank-band or sampled alternative was as damaging as the localized top-20 set for any model.
}
\label{tab:conditional-ablation-control}
\end{table*}

\begin{table*}[h]
\centering
\footnotesize
\setlength{\tabcolsep}{4pt}
\textit{Removing the selected source circuit}

\vspace{0.25em}

\begin{tabular}{lrrrr}
\toprule
Model & Removed budget & Source faithfulness & Removed edges & Max faithfulness \\
\midrule
GPT-2 & 1,342 & 0.862067 & 3,188 & 0.000433 \\
Llama 3.2 3B & 16,806 & 0.920909 & 42,230 & 0.000183 \\
Gemma 3 4B & 4,992 & 0.879058 & 12,570 & 0.000066 \\
Qwen 3 4B & 12,017 & 0.855521 & 28,099 & 0.000029 \\
\bottomrule
\end{tabular}

\vspace{0.75em}

\textit{Removing only the localized top-20 source edges}

\vspace{0.25em}

\begin{tabular}{lrrrr}
\toprule
Model & Removed budget & Source faithfulness & Removed edges & Max faithfulness \\
\midrule
GPT-2 & 20 & 0.862067 & 32 & 0.023524 \\
Llama 3.2 3B & 20 & 0.920909 & 24 & 0.033290 \\
Gemma 3 4B & 20 & 0.879058 & 22 & 0.056377 \\
Qwen 3 4B & 20 & 0.855521 & 36 & 0.016373 \\
\bottomrule
\end{tabular}

\caption{Shadow-circuit rediscovery after removing the selected source circuit (top) or only its localized top-20 source edges (bottom). The selected circuit is the smallest saved source budget whose full-model sweep reached \texttt{faithfulness\_mean} $\geq 0.85$; the top-20 rows use the same source circuit's faithfulness because the removed edges are drawn from that source ranking. Removed budget counts collapsed edges, removed edges counts underlying graph edges, and max faithfulness is the maximum validation faithfulness achieved by any budget in the reduced-graph sweep.}
\label{tab:shadow-rediscovery}
\end{table*}


\subsection{Comparison with random initializations}
To test the robustness and reliability of our circuit we repeated the discovery across three different initialization with 3 different random seeds (0, 1, 42), the discovery has been done using also a different discover subsets to ensure that the circuit found does not rely on semantic cues specific of the minimal pairs used to find the circuit. Our results show how the three different runs lead to the same faithfulness at the same budget (Figure\ref{fig:budget_analysis}). To further investigate whether the circuits found with different initialization overlap we compute both the IoU and the mean edge overlap across seeds. In general, the budgets the overlap is very high, always contained in [0.86, 1.0] for the mean edge overlap and between [0.8, 1.0] for IoU.

\subsection{Comparison with random circuits}
\begin{figure*}
    \centering
    \includegraphics[width=1\linewidth]{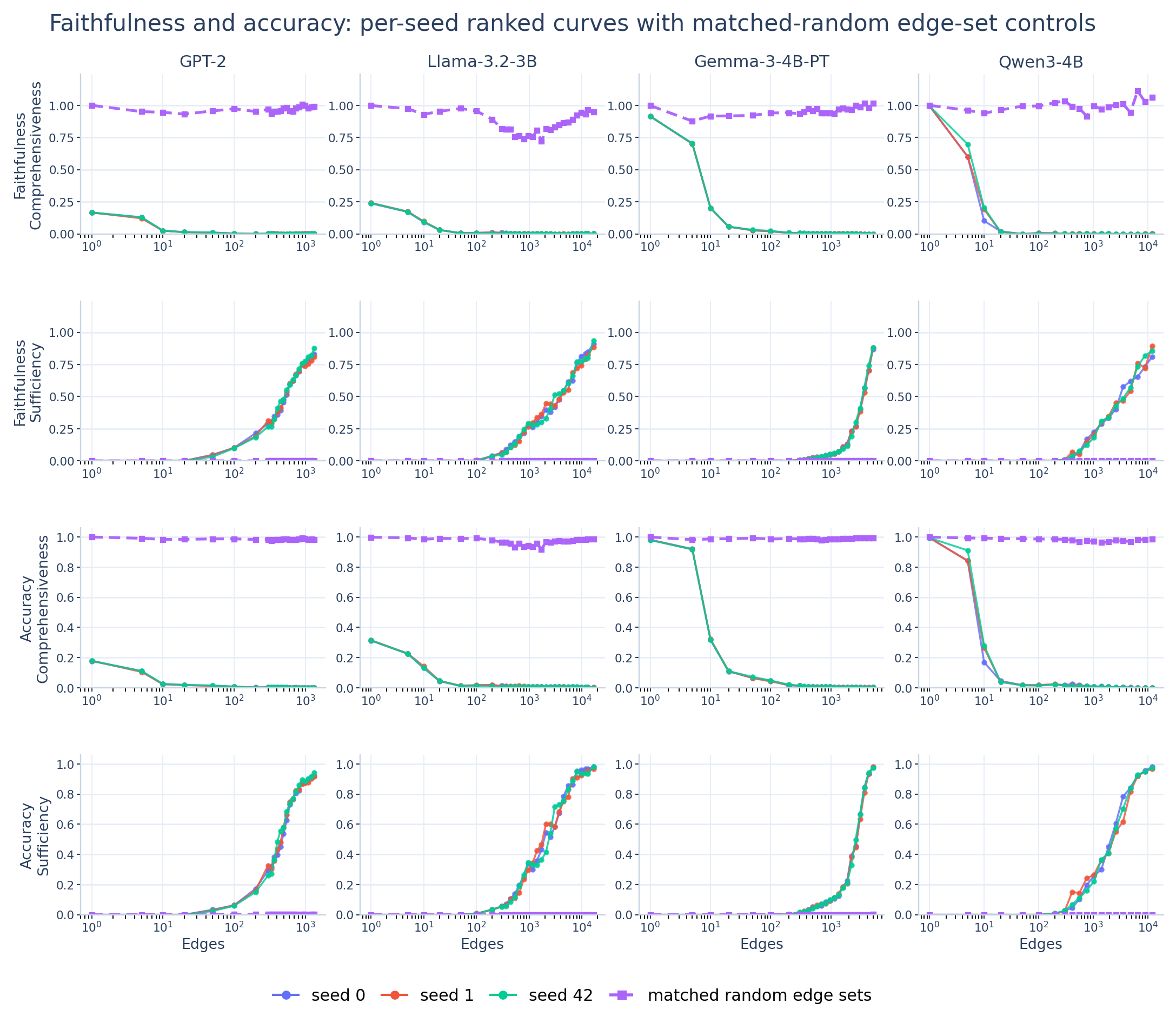}
    \caption{Faithfulness and accuracy comparisons between ranked circuits and matched-random edge sets. For each model, comprehensiveness is measured after cumulatively ablating the top-\(k\) ranked collapsed edge groups or matched-random edge groups, while sufficiency is measured by keeping only those edge groups. Random edge sets are matched
  to the ranked edges by edge type and layer where possible. Across both metrics, ranked edges recover model behavior when kept and strongly degrade behavior when ablated, whereas matched-random edge sets have little effect under ablation and fail to recover the task when kept alone.}
    \label{fig:budget_analysis}
\end{figure*}

To test whether the recovered circuits are meaningful rather than a consequence of a high budget alone, we compare them against randomly sampled edge sets of the same size. We perform this comparison across the full budget sweep for both sufficiency and necessity. For sufficiency, we retain only the randomly sampled edges and measure whether they recover the task behavior. For necessity, we ablate randomly sampled edges and measure whether they degrade performance to the same extent as ablating the discovered circuit. Across models, random circuits do not restore meaningful task performance when retained, nor do they produce comparable degradation when ablated (Figure~\ref{fig:budget_analysis}). At the budget where the discovered circuit first reaches \(85\%\) faithfulness, the faithfulness gap between the discovered circuit and the random baseline is 87.6\%, 93.5\%, 87.9\%, and 85.6\% for GPT-2, Llama 3.2 3B, Gemma 3 4B, and Qwen 3 4B, respectively.


\begin{table*}[t]
\centering
\scriptsize
\setlength{\tabcolsep}{3pt}
\resizebox{\textwidth}{!}{%
\begin{tabular}{lrrrrrrrrrrr}
\toprule
Model & Retained & Eval. & Orig. budget & Orig. Faith. & Orig. Acc. &
Task-2 budget & Task-2 Faith. & Task-2 Acc. & Best budget & Best Faith. &
Best Acc. \\
\midrule
GPT-2 & 2,416 & 1,916 & 1,342 & 0.526 & 0.693 & 995 & 0.886 & 0.864 & 1,342 & 1.033 & 0.926 \\
Llama 3.2 3B & 926 & 426 & 16,806 & 0.910 & 0.892 & 1,358 & 0.968 & 0.810 & 6,143 & 1.538 & 0.951 \\
Gemma 3 4B & 4,233 & 3,733 & 4,992 & 0.791 & 0.827 & 4,992 & 0.877 & 0.868 & 6,949 & 1.016 & 0.922 \\
Qwen 3 4B & 1,378 & 878 & 12,017 & 0.596 & 0.793 & 8,836 & 0.902 & 0.924 & 16,344 & 1.019 & 0.959 \\
\bottomrule
\end{tabular}%
}
\caption{Independent circuit discovery on Task 2, using the original prefixes with named-entity targets. Orig. 85\% evaluates the original common-noun animacy circuit on the Task 2 validation split. Task-2 85\% is the smallest independently discovered Task 2 circuit whose validation faithfulness reaches 0.85. Best is the highest-faithfulness row in the Task 2 budget sweep.}
\label{tab:task2-circuit-discovery}
\end{table*}

\section{Construction of the transfer datasets}
\label{app:transfer_tasks}
We construct three transfer datasets by independently varying the prefix distribution and the target set used to define the animacy-completion metric. This lets us separate two possible forms of transfer: transfer to new syntactic/lexical contexts with the same common-noun target vocabulary, and transfer to a different target vocabulary while keeping the same underlying animate-versus-inanimate distinction.

\paragraph{Setup 1: BLiMP prefixes with original common-noun targets}
For the first transfer setup, we replace our original semantic-prefix dataset with the BLiMP \texttt{animate\_subject\_passive} minimal-pair dataset. We use the \texttt{one\_prefix\_prefix} field from BLiMP, which gives prefixes such as "Amanda was respected by some" and "Lisa was kissed by the". These prefixes already end immediately before the target noun, so we can evaluate them with the same common-noun animate and inanimate target sets used in the original circuit-discovery runs. We keep all rows for which BLiMP marks the one-prefix and simple language-modeling evaluations as valid. This gives 1,000 BLiMP prefixes. For each prefix, we compute the same average animate-minus-inanimate logit difference used in the original task, but now on BLiMP passive-subject contexts rather than on our generated semantic-prefix contexts.

\paragraph{Setup 2: Original prefixes with named-entity targets}
For the second transfer setup, we keep the original clean/corrupt prefix pairs but replace the common-noun targets with named entities. The named-entity target set contains person names as animate targets and non-person named entities as inanimate targets, including locations, organizations, products, and works. Since named entities do not naturally follow the original "by the" template, we remove the terminal article from each prefix. Thus a prefix such as "The manuscript was annotated by the" is converted to "The manuscript was annotated by". We apply this transformation to both the clean and corrupt prefixes.

Many named entities are multi-token strings, so for these transfer experiments we use the first token of each entity as the target token. Duplicate first-token targets are collapsed. The final named-entity target vocabulary contains $\sim$100 animate first-token targets and $\sim$100 inanimate first-token targets. After scoring the truncated prefixes with this named-entity metric, we retain only examples for which the full model prefers the animate targets on the clean prefix and the inanimate targets on the corrupt prefix. As in the main discovery setting, we then sample 500 examples for circuit discovery and use the remaining retained examples for validation.

\paragraph{Setup 3: BLiMP prefixes with named-entity targets}
The third transfer setup combines both distribution shifts. We start from the same BLiMP \texttt{animate\_subject\_passive} prefixes used in the first transfer task, but evaluate them with the named-entity target sets used in the second transfer task. Because named entities require the bare \textit{by} template, we strip the final determiner from each BLiMP prefix: prefixes ending in "by some" or "by the" are truncated to end in "by". For example, "Amanda was respected by some" becomes "Amanda was respected by". We then evaluate the resulting 1,000 truncated BLiMP prefixes using the same first-token named-entity target metric.

Together, these three datasets test whether the discovered circuit is tied only to the original prefix distribution and common-noun answer vocabulary, or whether it captures a more general animate-versus-inanimate completion behavior. Setup 1 changes the prefix distribution while keeping the original target vocabulary fixed; setup 2 changes the target vocabulary while keeping the original prefix source fixed; and setup 3 changes both at once.

\begin{table*}[t]
\centering
\scriptsize
\setlength{\tabcolsep}{4pt}
\resizebox{\textwidth}{!}{%
\begin{tabular}{lrrrrrrr}
\toprule
Model & Orig. budget & Setup-2 budget & Shared edges & Orig. overlap &
Setup-2 overlap & IoU & Orig.-budget IoU \\
\midrule
GPT-2 & 1,342 & 995 & 687 & 0.512 & 0.690 & 0.416 & 0.457 \\
Llama 3.2 3B & 16,806 & 1,358 & 1,272 & 0.076 & 0.937 & 0.075 & 0.452 \\
Gemma 3 4B & 4,992 & 4,992 & 4,041 & 0.809 & 0.809 & 0.680 & 0.680 \\
Qwen 3 4B & 12,017 & 8,836 & 5,563 & 0.463 & 0.630 & 0.364 & 0.389 \\
\bottomrule
\end{tabular}%
}
\caption{Collapsed-edge overlap between the original 85\%-faithfulness circuit and the independently discovered setup 2 circuit. Setup-2 overlap is the fraction of the Setup-2 85\% circuit that also appears in the original 85\% circuit. Orig. overlap is the corresponding fraction of the original circuit. Orig.-budget IoU compares the original circuit to the setup 2 ranking at the original circuit budget.}
\label{tab:task2-circuit-overlap}
\end{table*}

\section{Setup 2 Circuit Discovery}
\label{app:task2-circuit-discovery}

Here, we report the circuit-discovery results for setup 2, which keeps our original clean/corrupt prefix distribution but replaces the common-noun target vocabulary with named entities.

As in the transfer construction for setup 2, prefixes of the form \textit{The [patient] was [verb] by the} become \textit{The [patient] was [verb] by}. This makes the prompts compatible with named entities, which do not naturally follow the determiner \textit{the}. The named-entity target file contains $\sim$100 animate person names and $\sim$100 inanimate or non-person named entities. Since many named entities are multi-token strings, we use the first tokenizer token of each entity and collapse duplicate first-token targets.

We then score the truncated original prefix pairs with the same average logit-difference metric used in the main experiments. For each model, we retain only the examples on which the full model satisfies the two sign conditions: animate preference on the clean prefix and inanimate preference on the corrupt prefix. From this retained set, we sample 500 high-margin examples for EAP-IG discovery and evaluate the resulting ranked circuits on the remaining examples. Table~\ref{tab:task2-circuit-discovery} summarizes the circuit discovery and transfer results.

The transferred original circuits only partially recover the named-entity behavior. Llama 3.2 3B is the strongest case, reaching faithfulness 0.910 and accuracy 0.892, but GPT-2 and Qwen 3 4B drop much more sharply, with faithfulness 0.526 and 0.596. Gemma 3 4B falls between these cases, with faithfulness 0.791 and accuracy 0.827.

Running EAP-IG directly on setup 2 recovers high-faithfulness circuits for all four models. The first circuits that reach the 0.85 faithfulness threshold have faithfulness between 0.877 and 0.968. In GPT-2 and Qwen 3 4B, direct setup 2 discovery also improves accuracy substantially relative to transferring the original circuit, from 0.693 to 0.864 for GPT-2 and from 0.793 to 0.924 for Qwen 3 4B. Llama 3.2 3B shows a different pattern: the threshold circuit is very faithful but still has lower accuracy, suggesting that it recovers the magnitude of the full-model logit-difference effect more reliably than the sign condition on every validation example. At larger budgets, the best highest-faithfulness rows reach accuracies of 0.926 for GPT-2, 0.951 for Llama 3.2 3B, 0.922 for Gemma 3 4B, and 0.959 for Qwen 3 4B. As in the main experiments, faithfulness can exceed 1 because it is normalized by the full-model clean--corrupt effect; a circuit can therefore over-recover that effect on the validation split. Table~\ref{tab:task2-circuit-overlap} reports the collapsed-edge overlap between the independently discovered setup 2 circuits and the original animacy circuits.

The independently discovered setup 2 circuits are related to the original animacy circuits, but they are not identical copies. Gemma 3 4B has the largest exact overlap: 4,041 of its 4,992 Task 2 threshold edges are also in the original circuit, giving IoU 0.680. GPT-2 and Qwen 3 4B show more moderate overlap, with IoU 0.416 and 0.364. Llama 3.2 3B has low IoU at the threshold budget because the setup 2 threshold circuit is much smaller than the original one; nevertheless, 1,272 of its 1,358 Task 2 threshold edges are contained in the original circuit. Comparing the setup 2 ranking at the original circuit size raises the Llama IoU to 0.452.

Overall, setup 2 supports a mixed conclusion. The original animacy circuits do carry useful information from common-noun targets to named-entity targets, and the independently discovered setup 2 circuits reuse many of the same collapsed edges. At the same time, direct discovery on the named-entity setup gives substantially stronger faithfulness and, in most cases, stronger accuracy than transferring the original circuit unchanged. This suggests that the models use a partly shared animacy mechanism across target vocabularies, but that the edge ranking and circuit size remain sensitive to the target set used to define the metric.

\clearpage
\onecolumn
\section{Necessary core subcircuits}
\begin{figure*}[!ht]
    \centering
    \includegraphics[width=\linewidth]{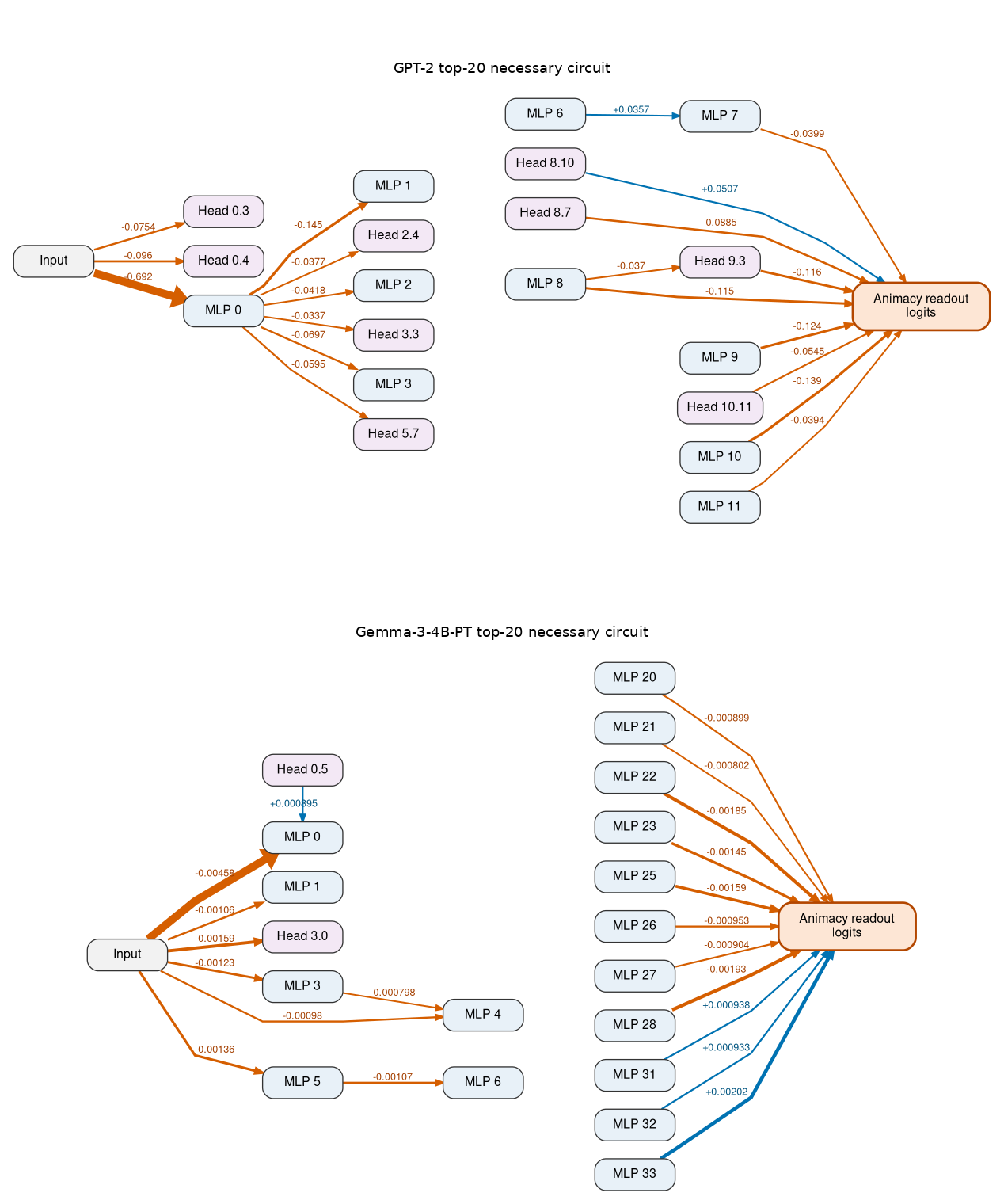}
    \caption{Visualization of the core necessary subcircuit for GPT-2 small and Gemma 3}
    \label{fig:fig:circuit1}
\end{figure*}

\begin{figure*}
    \centering
    \includegraphics[width=\linewidth]{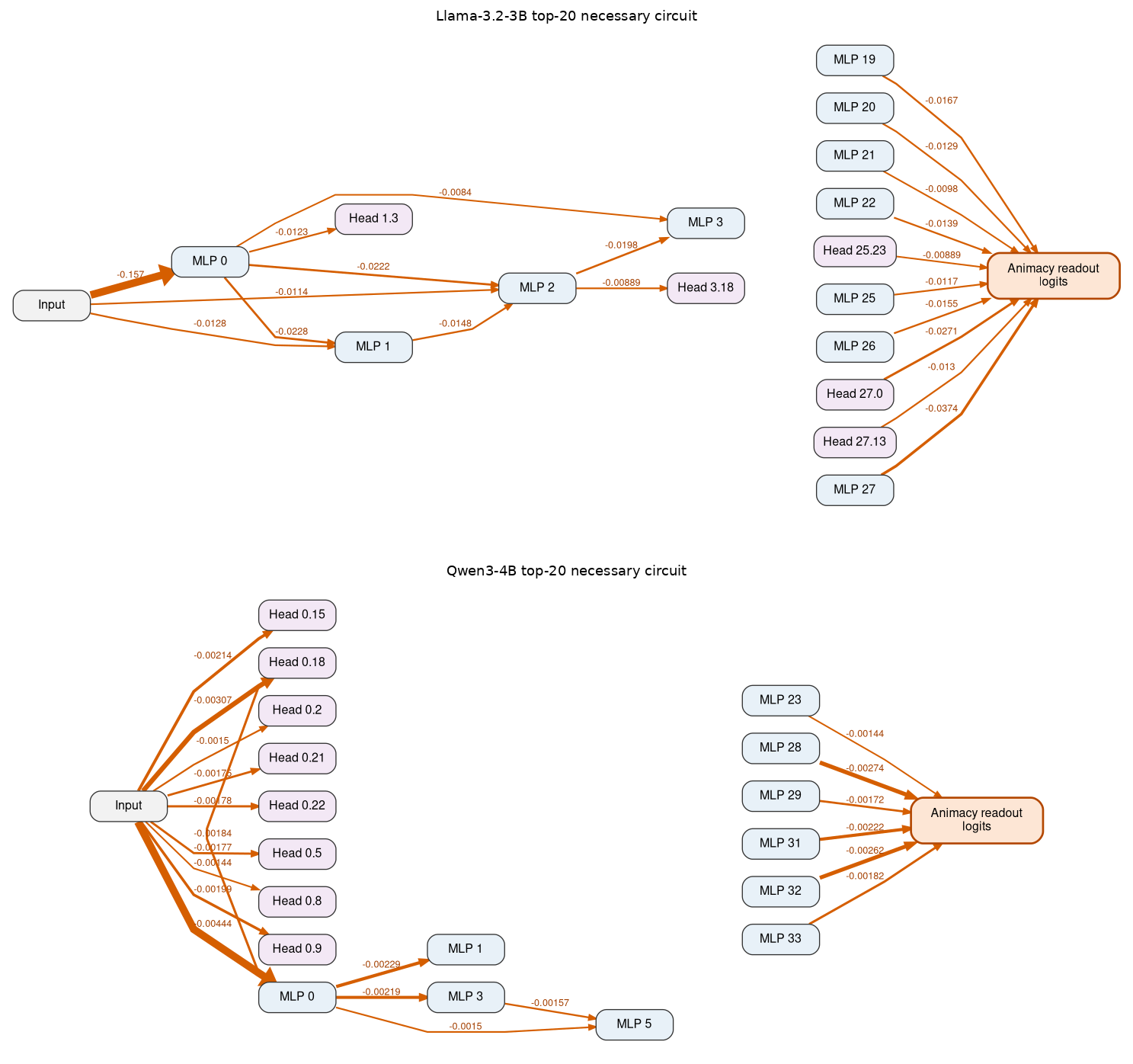}
    \caption{Visualization of the core necessary subcircuit for Llama 3.2 and Qwen 3}
    \label{fig:circuit2}
\end{figure*}

\clearpage
\section{Induced score of necessary circuit components}

\begin{figure*}[!ht]
    \centering
    \includegraphics[width=0.8\linewidth]{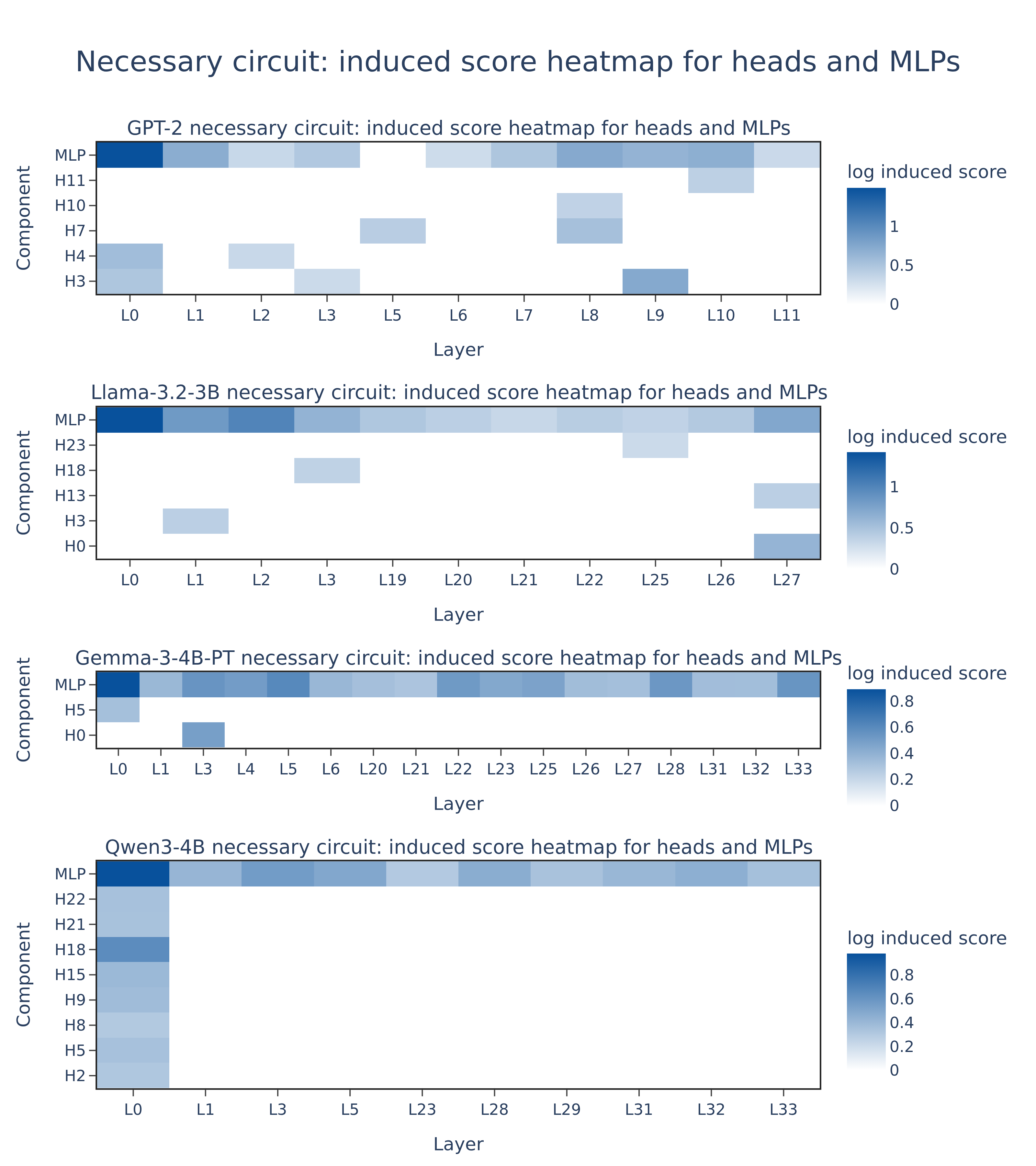}
    \caption{Induced score of necessary circuit components}
    \label{fig:induced_score_necessary}
\end{figure*}

\clearpage
\section{Logits contribution of necessary circuit components}

\begin{figure*}[h]
    \centering
    \includegraphics[width=1\linewidth]{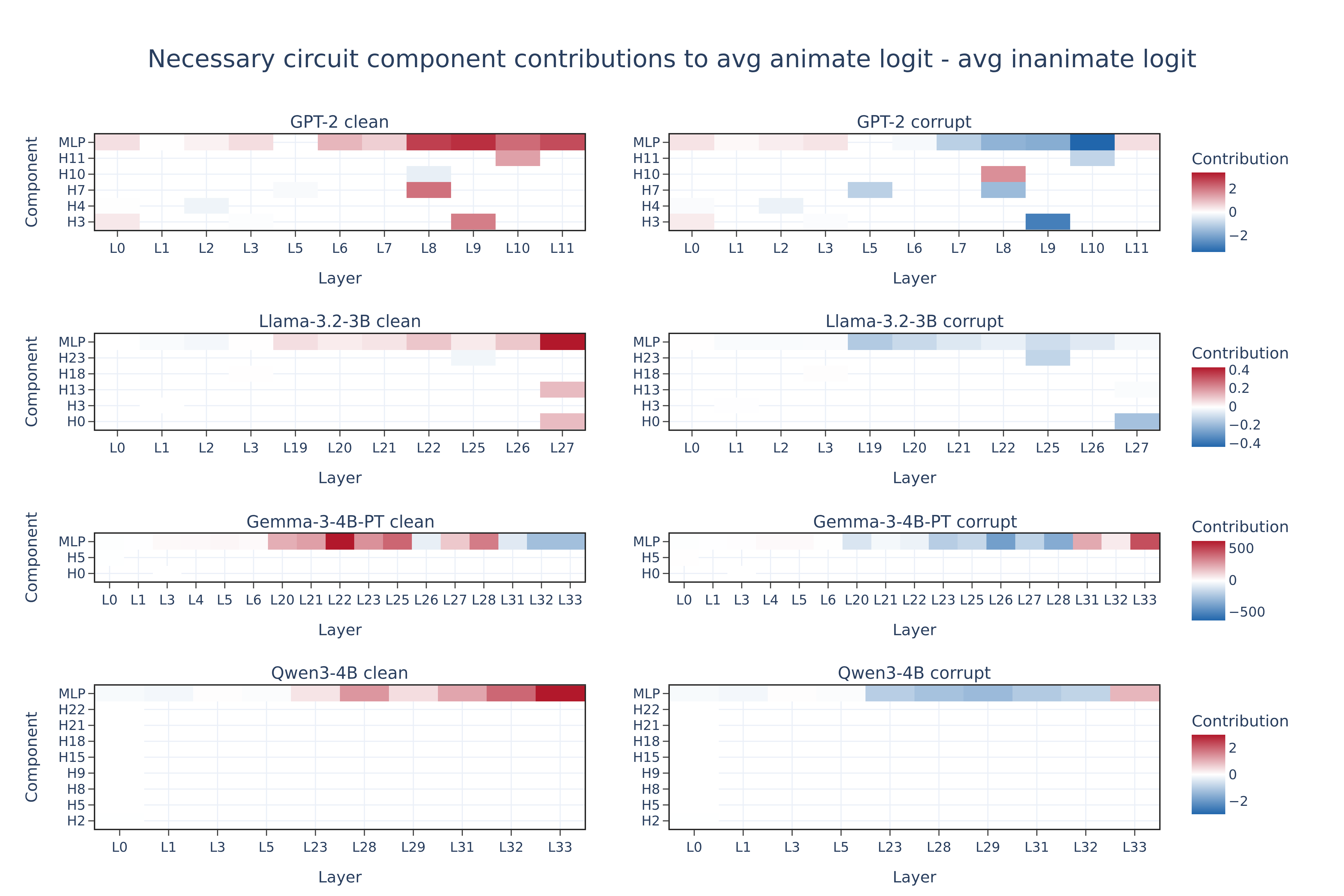}
    \caption{Direct logit contributions of necessary-circuit components to the animate--inanimate target preference. For each model, we plot the contribution of the selected necessary-core components to the average animate logit minus the average inanimate logit, separately for clean and corrupt prefixes. Red indicates components that increase the animate-over-inanimate preference, while blue indicates components that decrease it. Across models, later MLPs contribute more directly to the final logit difference than early components, suggesting a distinction between components that participate in the circuit and components that directly write the animacy preference to the logits.}
    \label{fig:logit_contribution}
\end{figure*}


\clearpage
\twocolumn
\section{Attention patterns of the necessary circuit}

\begin{figure*}[h]
    \centering
    \includegraphics[width=\linewidth]{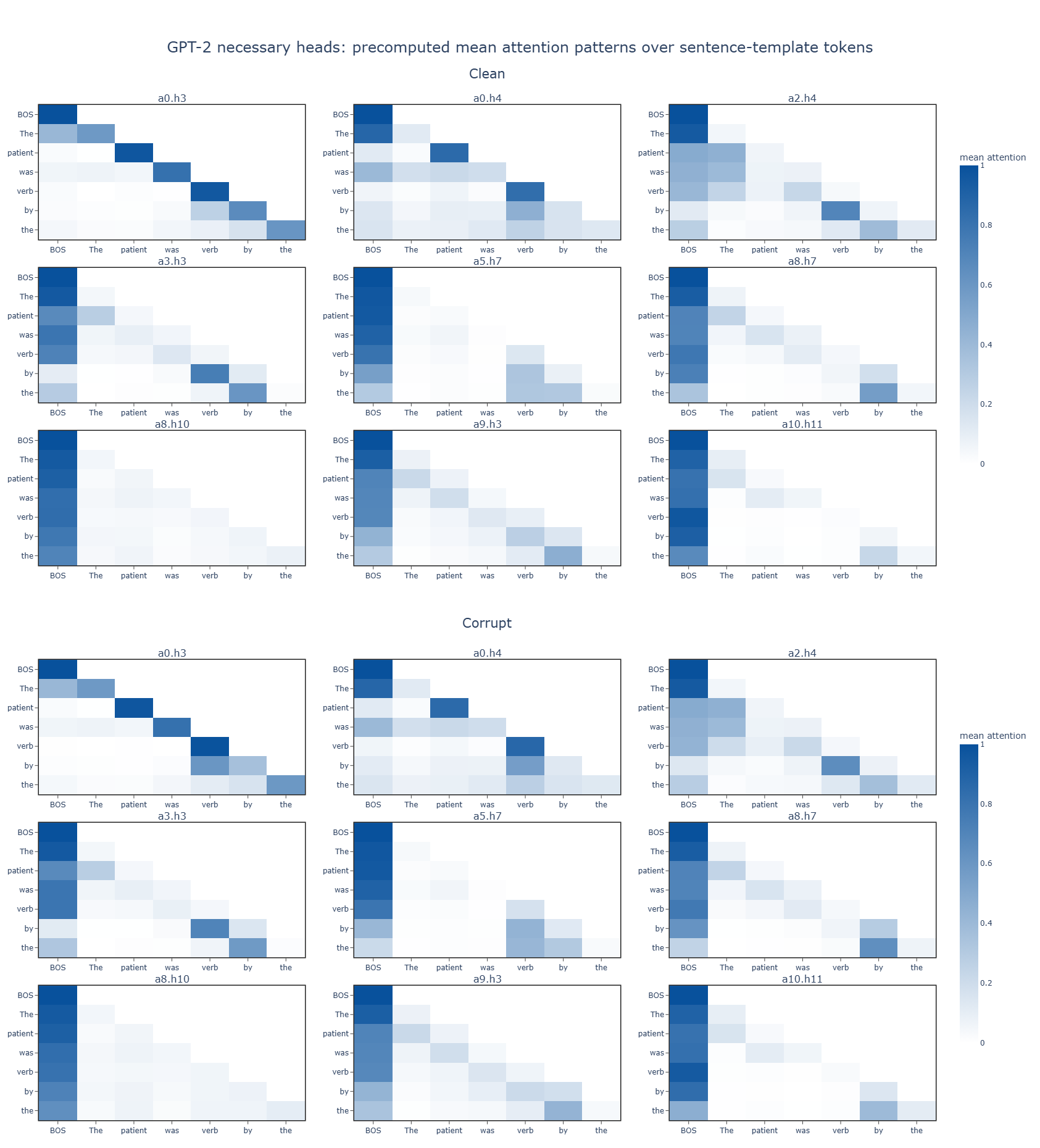}
    \caption{Attention head pattern of important heads in the GPT-2 small necessary circuit}
    \label{fig:attention_patterns_gpt2}
\end{figure*}

\paragraph{GPT-2 Small.}
The necessary attention heads in GPT-2 (Figure\ref{fig:attention_patterns_gpt2}) reveal a largely positional routing structure. Several later heads, including a5.h7, a8.h7, a8.h10, a9.h3, and a10.h11, show strong BOS-directed attention, a phenomenon known as attention sink \cite{ICLR2024_5e5fd18f} frequently observed in decoder-only language models \citep{gu2025attention}. Earlier heads, such as a0.h4, a2.h4, and a3.h3, show more local patterns consistent with information moving from the verb region toward the passive marker \textit{by}. This suggests that GPT-2 may use a combination of BOS-mediated routing and local verb/\textit{by} interactions to support the animacy-sensitive prediction.

\begin{figure*}
    \centering
    \includegraphics[width=1\linewidth]{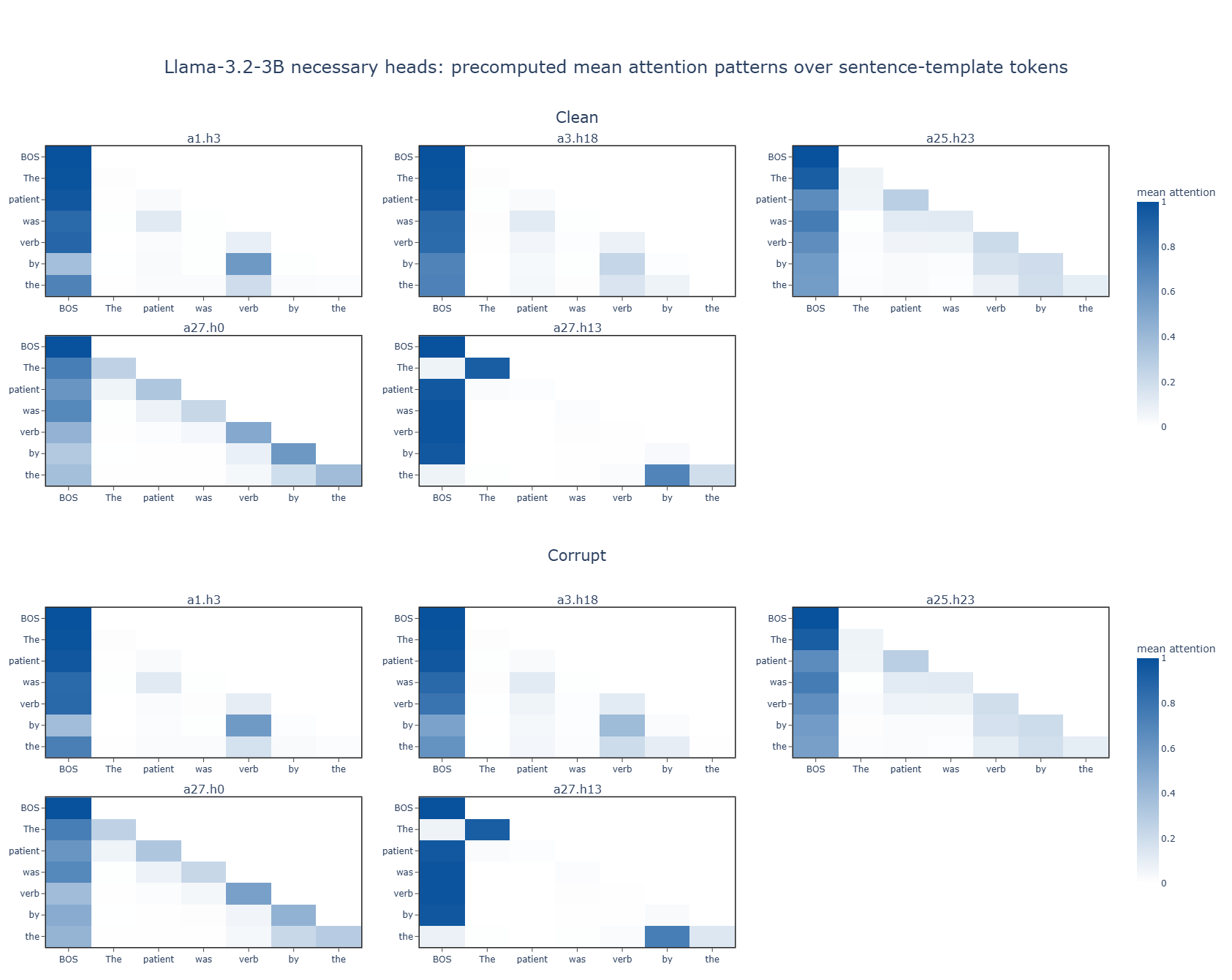}
    \caption{Attention head pattern of important heads in the Llama necessary circuit}
    \label{fig:attention_patterns_llama}
\end{figure*}

\paragraph{Llama 3.2 3B.}
The necessary Llama heads (Figure\ref{fig:attention_patterns_llama}) show the clearest positional relay. In early heads a1.h3 and a3.h18, the passive marker \textit{by} attends strongly to the verb position. Later, head a27.h13 shows the following determiner \textit{the} attending primarily to \textit{by}, with comparatively little attention to BOS. This supports a structural hypothesis in which verb-related information is first made available at the passive marker and then carried forward to the beginning of the upcoming agent phrase through a \textit{verb} $\rightarrow$ \textit{by} $\rightarrow$ \textit{the} pathway.

\begin{figure*}
    \centering
    \includegraphics[width=1\linewidth]{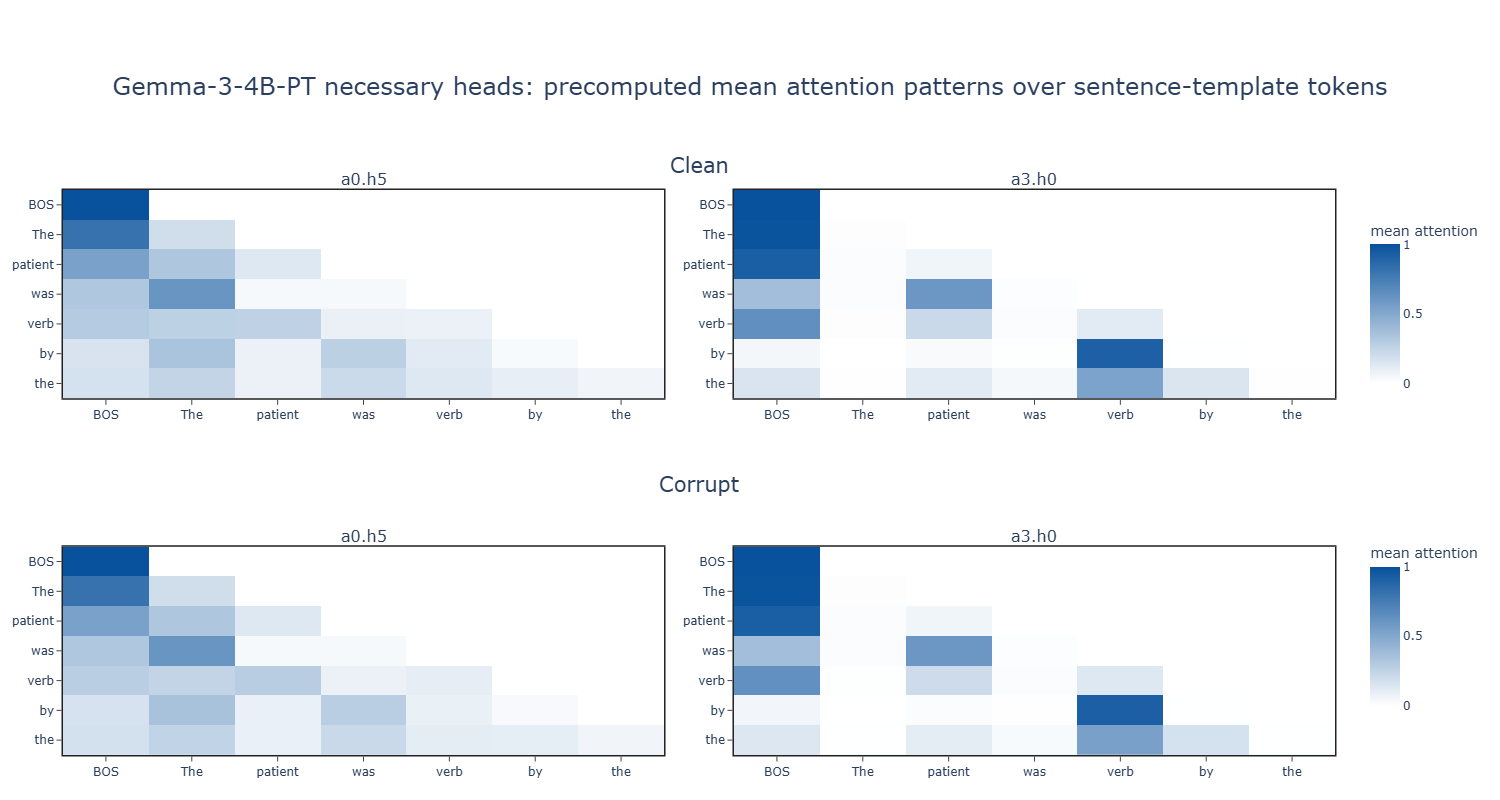}
    \caption{Attention head pattern of important heads in the Gemma necessary circuit}
    \label{fig:attention_patterns_gemma}
\end{figure*}

\paragraph{Gemma 3 4B.}
The necessary attention component for Gemma (Figure\ref{fig:attention_patterns_gemma}) is considerably smaller, containing only two heads. Head a0.h5 shows a broad early-layer pattern, combining attention to BOS with diffuse attention over previous sentence positions, and is difficult to assign to a specific functional role. Head a3.h0 is more interpretable: the passive marker \textit{by} attends strongly to the verb position, and the final determiner shows weaker attention to the verb/\textit{by} region. This is compatible with verb-region information being made available around the passive marker, although Gemma does not show the clearer later \textit{by} $\rightarrow$ \textit{the} relay observed in Llama.

\begin{figure*}
    \centering
    \includegraphics[width=1\linewidth]{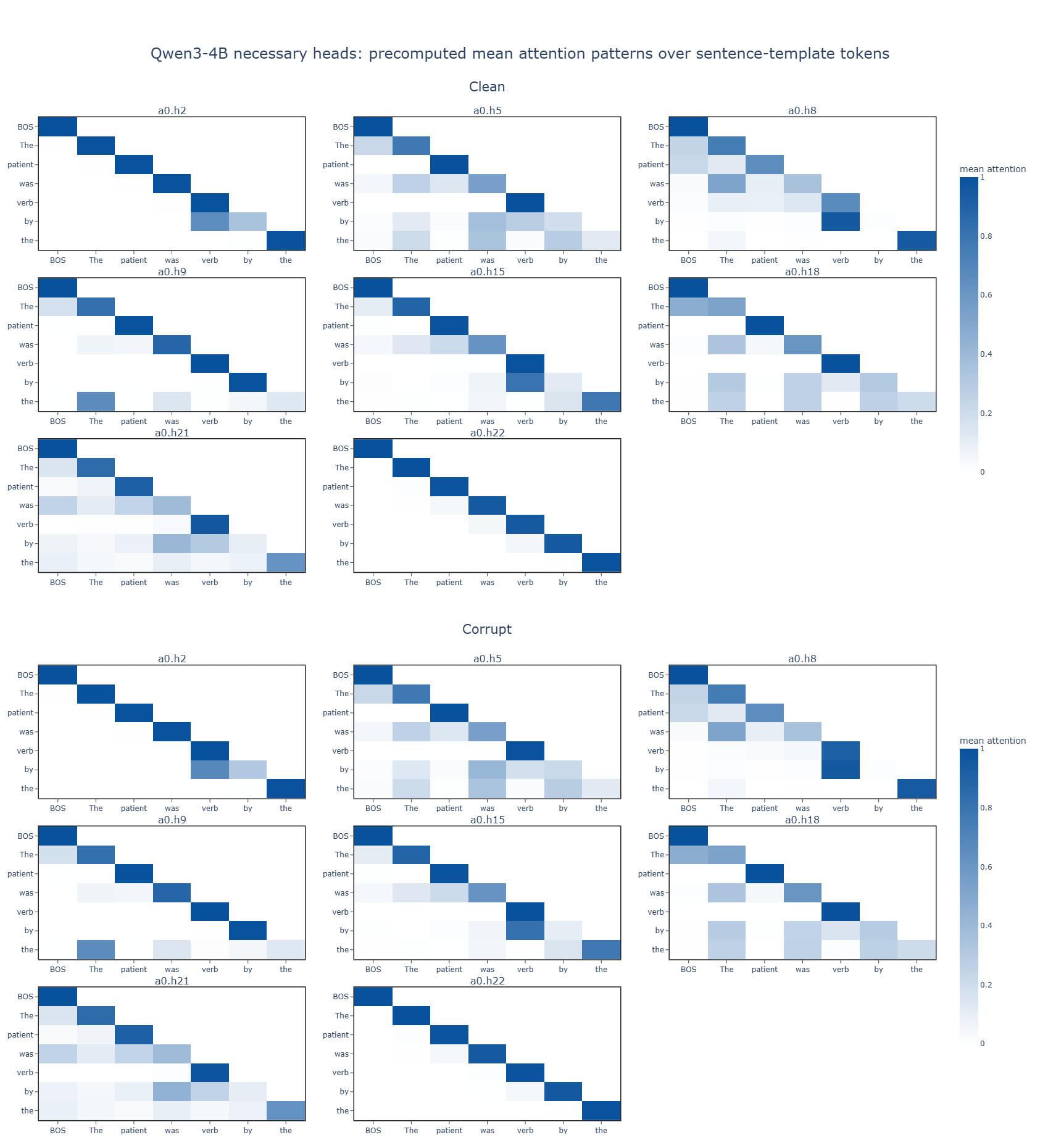}
    \caption{Attention head pattern of important heads in the Qwen necessary circuit}
    \label{fig:attention_patterns_qwen}
\end{figure*}

\paragraph{Qwen 3 4B.}
The necessary attention heads in Qwen (Figure\ref{fig:attention_patterns_qwen}) are harder to interpret. Most selected heads are in layer 0 and show largely diagonal or near-diagonal attention, suggesting generic token-local processing rather than a specialized animacy pathway. A few heads show weak structure around the verb and passive marker, such as attention from \textit{by} to the verb region in a0.h2, a0.h5, and a0.h15, but this pattern is not as isolated as in Llama or Gemma. Overall, Qwen's necessary attention component appears dominated by early positional or token-local operations, with no clear multi-step \textit{verb} $\rightarrow$ \textit{by} $\rightarrow$ \textit{the} pathway.

\clearpage
\section{Circuit Discovery on Model-Specific vs Shared Datasets}
\label{app:shared-discovery}

In the main experiments, each model's circuit was discovered on its own model-specific retained dataset: the subset of the common tokenizer-compatible
prompt-pair pool on which that model succeeds under the average animate--inanimate logit-difference metric. As an additional check, we also ran the same full-model EAP-IG discovery procedure on the smaller shared dataset. This shared dataset contains only prompt pairs retained across the four-model intersection, giving 3,076 retained examples for every model. In both settings, we sampled 500 high-margin examples for discovery and evaluated the resulting ranked circuits on the remaining retained examples. Table~\ref{tab:shared-discovery} summarizes circuit size, faithfulness, and accuracy for discovery on the model-specific and shared retained datasets.

We restrict this comparison to pure circuit discovery and the associated validation budget sweep. In particular, the shared-dataset runs were not followed by the diagnostic analyses used in the paper, such as random-edge controls, necessity ablations. 

Across all four models, discovery on the shared dataset still recovers circuits above the 0.85 faithfulness threshold. For GPT-2, the shared and model-specific runs reach the threshold at the same collapsed-edge budget, with nearly identical accuracy and slightly higher faithfulness on the shared validation set. Llama 3.2 3B and Gemma 3 4B also reach the threshold at the same budgets as in the model-specific runs, with small changes in faithfulness and accuracy. Qwen 3 4B shows the largest difference: the shared run reaches higher faithfulness (0.891 vs. 0.856) with a smaller collapsed-edge budget (8,836 vs. 12,017), although its validation accuracy is slightly lower.

As shown in Table~\ref{tab:shared-discovery-overlap}, GPT-2, Llama 3.2 3B, and Gemma 3 4B
all have mean edge overlap above 0.91 and IoU above 0.83 between the model-specific and shared circuits. Qwen 3 4B has lower IoU because the shared threshold circuit is smaller, but most of its shared circuit is still contained in the model-specific one: 8,594 of the 8,836 shared collapsed edges also appear in the model-specific threshold circuit. Thus, changing from the model-specific retained dataset to the stricter shared retained dataset mostly preserves the ranked circuit structure, rather than producing a separate high-faithfulness edge set.

These results suggest that the main EAP-IG discovery result is not an artifact of giving each model only its own most favorable retained examples. A stricter (and smaller) cross-model shared dataset still yields high-faithfulness circuits for all four models with a high intersection compared to the original ones.

\begin{table*}[t]
\centering
\small
\setlength{\tabcolsep}{4pt}
\begin{tabular}{llrrrrrrr}
\toprule
Model & Dataset & Retained & Eval. & Budget & Faith. & Acc. & Expanded & Nodes \\
\midrule
GPT-2 & Model-specific & 4,721 & 4,221 & 1,342 & 0.862 & 0.940 & 3,111 & 109 \\
GPT-2 & Shared & 3,076 & 2,576 & 1,342 & 0.882 & 0.939 & 3,088 & 109 \\
Llama 3.2 3B & Model-specific & 4,734 & 4,234 & 16,806 & 0.921 & 0.981 & 42,008 & 508 \\
Llama 3.2 3B & Shared & 3,076 & 2,576 & 16,806 & 0.906 & 0.974 & 42,002 & 503 \\
Gemma 3 4B & Model-specific & 4,371 & 3,871 & 4,992 & 0.879 & 0.973 & 7,751 & 156 \\
Gemma 3 4B & Shared & 3,076 & 2,576 & 4,992 & 0.868 & 0.988 & 7,792 & 156 \\
Qwen 3 4B & Model-specific & 4,300 & 3,800 & 12,017 & 0.856 & 0.976 & 27,433 & 668 \\
Qwen 3 4B & Shared & 3,076 & 2,576 & 8,836 & 0.891 & 0.967 & 19,568 & 560 \\
\bottomrule
\end{tabular}
\caption{Full-model EAP-IG discovery on model-specific and shared retained datasets. The reported circuit is the smallest saved collapsed-edge budget whose validation faithfulness reaches at least 0.85. Retained is the number of examples after dataset filtering; Eval. is the number of validation examples after removing the 500-example discovery sample. Faith. is mean validation faithfulness, Acc. is mean validation accuracy, Expanded is the number of underlying graph edges induced by the collapsed-edge budget, and Nodes is the number of induced graph nodes.}
\label{tab:shared-discovery}
\end{table*}

\begin{table*}[b]
\centering
\setlength{\tabcolsep}{5pt}
\begin{tabular}{lrrrrr}
\toprule
Model & $|C_{\mathrm{model}}|$ & $|C_{\mathrm{shared}}|$ &
$|C_{\mathrm{model}} \cap C_{\mathrm{shared}}|$ &
Mean edge overlap & IoU \\
\midrule
GPT-2 & 1,342 & 1,342 & 1,236 & 0.921 & 0.854 \\
Llama 3.2 3B & 16,806 & 16,806 & 15,307 & 0.911 & 0.836 \\
Gemma 3 4B & 4,992 & 4,992 & 4,751 & 0.952 & 0.908 \\
Qwen 3 4B & 12,017 & 8,836 & 8,594 & 0.844 & 0.701 \\
\bottomrule
\end{tabular}
\caption{Exact collapsed-edge overlap between the model-specific and shared dataset circuits from Table~\ref{tab:shared-discovery}. For each model, we compare the top collapsed edges up to the first saved budget reaching faithfulness $\geq 0.85$ in each dataset setting. Mean edge overlap is the average of the two directional overlaps, $\frac{1}{2}(|A \cap B|/|A| + |A \cap B|/|B|)$. IoU is the Jaccard similarity, $|A \cap B|/|A \cup B|$.}
\label{tab:shared-discovery-overlap}
\end{table*}

\end{document}